\documentclass[10pt,twocolumn,letterpaper]{article}

\usepackage{iccv}

\usepackage{times}
\usepackage{epsfig}
\usepackage{graphicx}
\usepackage{caption}
\usepackage{subcaption}
\usepackage{amsmath}
\usepackage{amssymb}
\usepackage{booktabs}
\usepackage{bm}
\usepackage{algorithm}
\usepackage{algorithmic}
\usepackage{listings}
\usepackage{url}

\iccvfinalcopy 

\usepackage[capitalize]{cleveref}
\crefname{section}{Sec.}{Secs.}
\Crefname{section}{Section}{Sections}
\Crefname{table}{Table}{Tables}
\crefname{table}{Tab.}{Tabs.}

\def\iccvPaperID{8770} 

\ificcvfinal\fi

\def\ptsd{Pix2Seq-$\mathcal{D}$\xspace}

\begin{document}

\title{A Generalist Framework for Panoptic Segmentation of Images and Videos}

\author{Ting Chen\thanks{$^\dagger$ Equal advising.}, Lala Li, Saurabh Saxena, Geoffrey Hinton$^\dagger$, David J. Fleet$^\dagger$\\
Google Deepmind\\
{\tt\small \{iamtingchen,lala,srbs,geoffhinton,davidfleet\}@google.com}
}
\renewcommand\footnotemark{}
\maketitle
\ificcvfinal\thispagestyle{empty}\fi

\begin{abstract}
Panoptic segmentation assigns semantic and instance ID labels to every pixel of an image. As permutations of instance IDs are also valid solutions, the task requires learning of high-dimensional one-to-many mapping. As a result, state-of-the-art approaches use customized architectures and task-specific loss functions. We formulate panoptic segmentation as a discrete data generation problem, without relying on inductive bias of the task. A diffusion model is proposed to model panoptic masks, with a simple architecture and generic loss function.
By simply adding past predictions as a conditioning signal, our method is capable of modeling video (in a streaming setting) and thereby learns to track object instances automatically. With extensive experiments, we demonstrate that our simple approach can perform competitively to state-of-the-art specialist methods in similar settings.
\end{abstract}
\footnote{Code at https://github.com/google-research/pix2seq}

 \section{Introduction}

Panoptic segmentation~\cite{kirillov2019panoptic} is a fundamental vision task that assigns semantic and instance labels for every pixel of an image.
The semantic labels describe the class of each pixel (e.g., sky, vertical), and the instance labels provides a unique ID for each instance in the image (to distinguish different instances of the same class).
The task is a combination of semantic segmentation and instance segmentation, providing rich semantic information about the scene.

\begin{figure}[t!]
    \centering
    \includegraphics[width=\linewidth]{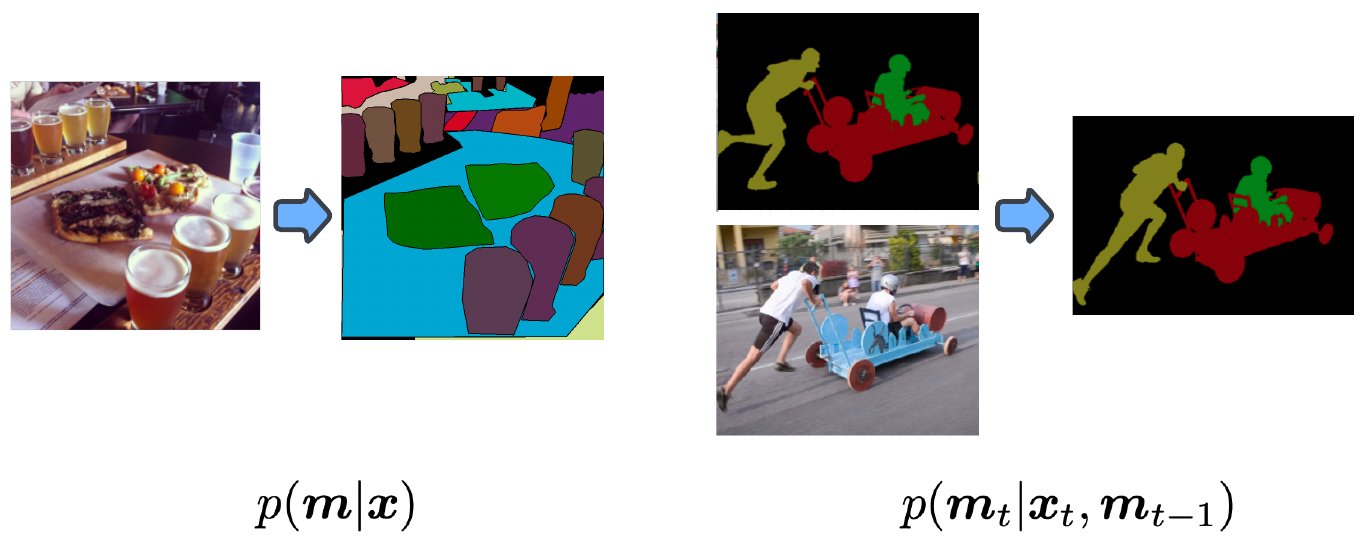}
    \vspace*{-0.5cm}
    \caption{We formulate panoptic segmentation as a conditional discrete mask ($\bm m$) generation problem for images (left) and videos (right), using a Bit Diffusion generative model~\cite{chen2022analog}.}
\label{fig:formulation}
    \vspace{-1em}
\end{figure}

While the class categories of semantic labels are fixed a priori, the instance IDs assigned to objects in an image can be permuted without affecting the instances identified.
For example, swapping instance IDs of two cars would not affect the outcome. Thus, a neural network trained to predict instance IDs should be able to learn a one-to-many mapping, from a single image to multiple instance ID assignments. The learning of one-to-many mappings is challenging and traditional approaches usually leverage a pipeline of multiple stages involving object detection, segmentation, merging multiple predictions~\cite{kirillov2019panoptic,li2019attention,xiong2019upsnet,liu2019end,cheng2020panoptic}. Recently, end-to-end methods~\cite{wang2021max,cheng2021per,zhang2021k,cheng2021masked,li2022panoptic,yu2022cmt,yu2022kmeans,li2022mask} have been proposed,  based on a differentiable bipartite graph matching~\cite{carion2020end}; this effectively converts a one-to-many mapping into a one-to-one mapping based on the identified matching. However, such methods still require customized architectures and sophisticated loss functions with built-in inductive bias for the panoptic segmentation task. 

Eshewing task-specific 
architectures and loss functions, 
recent generalist vision models, such as Pix2Seq~\cite{chen2021pix2seq,chen2022unified}, OFA~\cite{wang2022unifying}, UViM~\cite{UViM}, and Unified I/O~\cite{unifiedIO}, advocate a generic, task-agnostic framework,
generalizing across multiple tasks while being much simpler than previous models.
For instance, Pix2Seq~\cite{chen2021pix2seq,chen2022unified} formulates a set of core vision tasks in terms of the generation of semantically meaningful sequences conditioned on an image, and they train a single autoregressive model based on Transformers~\cite{vaswani2017attention}.

Following the same philosophy, we formulate the panoptic segmentation task as a conditional discrete data generation problem, depicted in Figure~\ref{fig:formulation}. We learn a generative model for panoptic masks, treated as an array of discrete tokens, conditioned on an input image. 
One can also apply the model to video data (in an online/streaming setting), by simply including predictions from past frames as an additional conditioning signal. 
In doing so, the model can then learn to track and segment objects automatically.

Generative modeling for panoptic segmentation is very challenging as the panoptic masks are discrete/categorical and can be very large.
To generate a 512$\times$1024 panoptic mask,
for example, the model has to produce more than 1M discrete tokens (of semantic and instance labels).
This is expensive for auto-regressive models as they are inherently sequential, scaling  poorly with the size of data input. Diffusion models~\cite{sohl2015deep,ho2020denoising,song2020denoising,song2021scorebased} are better at handling high dimension data but they are most commonly applied to continuous rather than discrete domains. By representing discrete data with analog bits~\cite{chen2022analog} we show that one can train a diffusion model on large panoptic masks directly, without the need to also learn an intermediate latent space.

In what follows, we introduce our diffusion-based model for panoptic segmentation, and then describe extensive experiments on both image and video datasets. In doing so we demonstrate that the proposed method performs competitively to state-of-the-art methods in similar settings, proving a simple and generic approach to panoptic segmentation.

\begin{figure*}[t!]
    \centering
    \includegraphics[width=.8\textwidth]{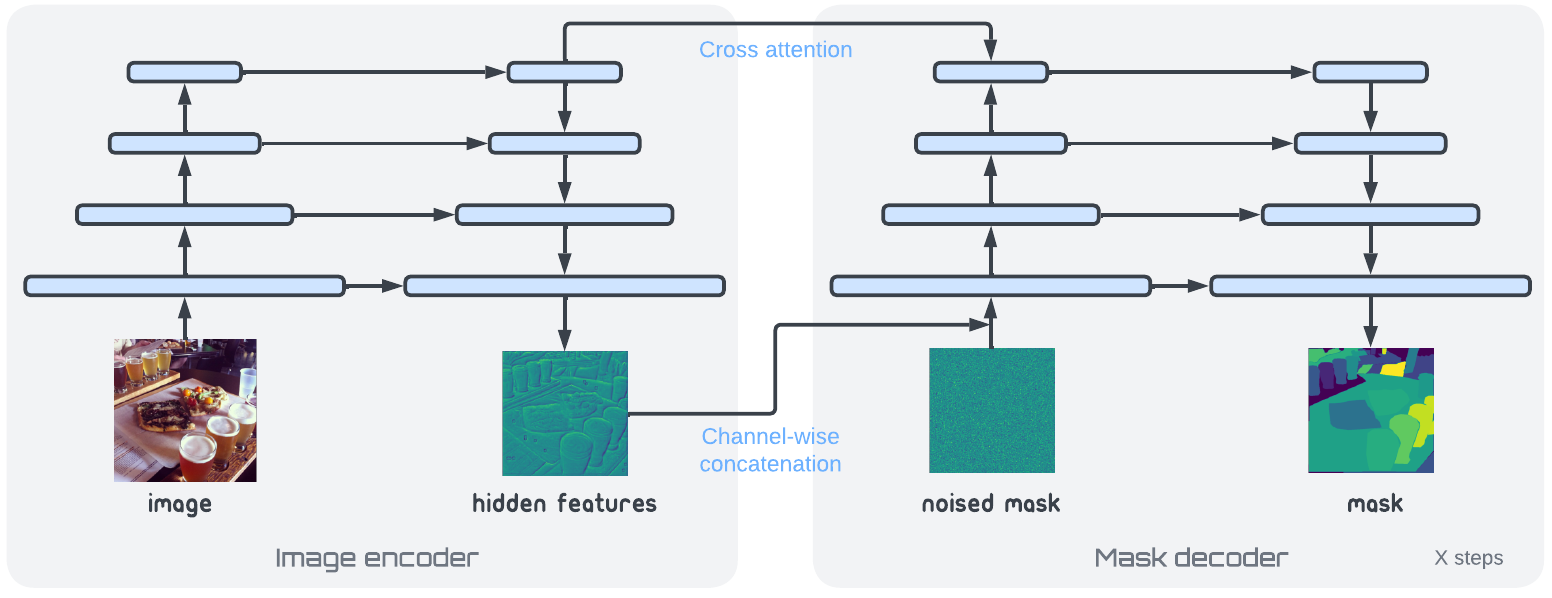}
    \vspace*{-0.2cm}
    \caption{The architecture for our panoptic mask generation framework. We separate the model into image encoder and mask decoder so that the iterative inference at test time only involves multiple passes over the decoder.}
    \label{fig:architecture}
    \vspace*{-0.2cm}
\end{figure*}

\section{Preliminaries}

\vspace*{-0.1cm}
\paragraph{Problem Formulation.}
Introduced in~\cite{kirillov2019panoptic}, panoptic segmentation masks can be expressed 
with two channels, $\bm m\in \mathbb{Z}^{H\times W\times 2}$.  The first represents the category/class label.  The second is the instance ID. 
Since instance IDs can be permuted without changing the underlying instances, we randomly assign integers in $[0, K]$  to instances every time an image is sampled during training.
$K$ is maximum number of instances allowed in any image (0 denotes the null label).

To solve the panoptic segmentation problem, we simply learn an image-conditional panoptic mask generation model by maximizing $\sum_i \log P(\bm m_i|\bm x_i)$, where $\bm m_i$ is a random categorical variable corresponding to the panoptic mask for image $\bm x_i$ in the training data. 
As mentioned above, panoptic masks may comprise hundreds of thousands or even millions of discrete tokens, making generative modeling very challenging, particularly for autoregressive models.

\vspace*{-0.2cm}
\paragraph{Diffusion Models with Analog Bits.}
Unlike autoregressive generative models, diffusion models are more effective with high dimension data ~\cite{sohl2015deep,ho2020denoising,song2020denoising,song2021scorebased} . Training entails learning a denoising network. During inference, the network generates target data in parallel, using far fewer iterations than the number of pixels.

In a nutshell, diffusion models learn a series of state transitions to transform noise $\bm \epsilon$ from a known noise distribution
into a data sample $\bm x_0$ from the data distribution $p(\bm x)$.
To learn this mapping, we first define a forward transition from
data $\bm x_0$ to a noisy sample $\bm x_t$ as follows,
\begin{equation}
\label{eq:diffusion_forward}
\bm x_t = \sqrt{\gamma(t)} ~\bm x_0 + \sqrt{1-\gamma(t)} ~\bm \epsilon,
\end{equation}
where $\bm \epsilon$ is drawn from standard normal
density, $t$ is from uniform density on [0,1], and $\gamma(t)$ is a monotonically decreasing function from 1 to 0.
During training, one learns a neural network $f(\bm x_t, t)$ to predict $\bm x_0$ (or $\bm \epsilon$) from $\bm x_t$, usually formulated as a denoising task with an $\ell_2$ loss:
\begin{equation}
\label{eq:diffusion_l2loss}
\mathcal{L}_{\bm x_0} =\mathbb{E}_{t\sim\mathcal{U}(0, T), \bm \epsilon\sim\mathcal{N}(\bm 0, \bm 1), \bm x_0}\|f(\bm x_t, t) - \bm x_0 \|^2.
\end{equation}
To generate samples from a learned model, it starts with a sample of noise, $\bm x_T$, and then follows a series of (reverse) state transitions $\bm x_T\rightarrow \bm x_{T-\Delta} \rightarrow  \cdots  \rightarrow \bm x_0$ by iteratively applying the denoising function $f$ with appropriate transition rules (such as those from DDPM~\cite{ho2020denoising} or DDIM~\cite{song2020denoising}).

Conventional diffusion models assume continuous data and Gaussian noise, and are not directly applicable to discrete data.
To model discrete data, Bit Diffusion~\cite{chen2022analog} first converts integers representing discrete tokens into bit strings, 
the bits of which are then cast as real numbers (a.k.a., analog bits) to which continuous diffusion models can be applied.
To draw samples, Bit Diffusion uses a conventional sampler from continuous diffusion, after which a final quantization step (simple thresholding) is used to obtain the categorical variables from
the generated analog bits.

\section{Method}

We formulate panoptic segmentation as a discrete data generation problem conditioned on pixels, similar to Pix2Seq~\cite{chen2021pix2seq,chen2022unified} but for dense prediction; hence we coin our approach \ptsd.
In what follows we first specify the architecture design, and then the training and inference algorithms based on Bit Diffusion.

\subsection{Architecture}
Diffusion model sampling is iterative, and hence one must run the forward pass of the network many times during inference. Therefore, as shown in Fig.\ \ref{fig:architecture}, we intentionally separate the network into two components: 1) an image encoder; and 2) a mask decoder.  
The former maps raw pixel data into high-level representation vectors, and then the mask decoder iteratively reads out the panoptic mask.

\vspace*{-0.2cm}
\paragraph{Pixel/image Encoder.} The encoder is a network that maps a raw image ${\bm x} \in \mathbb{R}^{H\times W\times 3}$ into a feature map in $\mathbb{R}^{H'\times W' \times d}$ where $H'$ and $W'$ are the height and width of the panoptic mask. The panoptic masks can be the same size or smaller than the original image. 
In this work, we follow~\cite{carion2020end,chen2021pix2seq} in using ResNet~\cite{he2016deep} followed by transformer encoder layers~\cite{vaswani2017attention} as the feature extractor. 
To make sure the output feature map has sufficient resolution, and includes features at different scales, inspired by U-Net~\cite{ho2020denoising,nichol2021improved,ronneberger2015u} and feature pyramid network~\cite{lin2017feature}, 
we use convolutions with bilateral connections and upsampling operations to merge features from different resolutions.
More sophisticated encoders are possible, to leverage recent advances in architecture designs~\cite{dosovitskiy2020image,liu2021swin,jaegle2021perceiver,zhang2022nested,khan2022transformers}, but this is not our main focus so we opt for simplicity.

\vspace*{-0.2cm}
\paragraph{Mask Decoder.} The decoder iteratively refines the panoptic mask during inference, conditioned on the image features. 
Specifically, the mask decoder is a TransUNet~\cite{chen2021transunet}.  It takes as input the concatenation of image feature map from encoder and a noisy mask (randomly initialized or from previous step), and outputs a refined prediction of the mask. One difference between our decoder and the standard U-Net architecture used for image generation and image-to-image translation \cite{ho2020denoising,nichol2021improved,Palette-SIGGRAPH2022} is that we use transformer decoder layers on the top of U-Net, with cross-attention layers to incorporate the encoded image features (before upsampling).

\subsection{Training Algorithm}

\begin{algorithm}[t]
\small
\caption{\small \ptsd training algorithm.
}
\label{alg:train}
\definecolor{codeblue}{rgb}{0.25,0.5,0.5}
\definecolor{codekw}{rgb}{0.85, 0.18, 0.50}
\lstset{
  backgroundcolor=\color{white},
  basicstyle=\fontsize{7.5pt}{7.5pt}\ttfamily\selectfont,
  columns=fullflexible,
  breaklines=true,
  captionpos=b,
  commentstyle=\fontsize{7.5pt}{7.5pt}\color{codeblue},
  keywordstyle=\fontsize{7.5pt}{7.5pt}\color{codekw},
  escapechar={|}, 
}
\begin{lstlisting}[language=python]
def train_loss(images, masks):
  """images: [b, h, w, 3], masks: [b, h', w', 2]."""
  
  # Encode image features.
  h = pixel_encoder(images)
  
  # Discrete masks to analog bits.
  m_bits = int2bit(masks).astype(float)
  m_bits = (m_bits * 2 - 1) * scale

  # Corrupt analog bits.
  t = uniform(0, 1)|~~~~~~~~~~~|# scalar.
  eps = normal(mean=0, std=1)  # same shape as m_bits.
  m_crpt = sqrt(|~~~~|gamma(t)) * m_bits + 
              sqrt(1 - gamma(t)) * eps

  # Predict and compute loss.
  m_logits, _ = mask_decoder(m_crpt, h, t)
  loss = cross_entropy(m_logits, masks)
  
  return loss.mean()
\end{lstlisting}
\end{algorithm}

\begin{algorithm}[t]
\small
\caption{\small \ptsd inference algorithm.
}
\label{alg:sample}
\definecolor{codeblue}{rgb}{0.25,0.5,0.5}
\definecolor{codekw}{rgb}{0.85, 0.18, 0.50}
\lstset{
  backgroundcolor=\color{white},
  basicstyle=\fontsize{7.5pt}{7.5pt}\ttfamily\selectfont,
  columns=fullflexible,
  breaklines=true,
  captionpos=b,
  commentstyle=\fontsize{7.5pt}{7.5pt}\color{codeblue},
  keywordstyle=\fontsize{7.5pt}{7.5pt}\color{codekw},
  escapechar={|}, 
}
\begin{lstlisting}[language=python]
def infer(images, steps=10, td=1.0):
  """images: [b, h, w, 3]."""
  
  # Encode image features.
  h = pixel_encoder(images)

  m_t = normal(mean=0, std=1)   # same shape as m_bits.
  for step in range(steps):
    # Get time for current and next states.
    t_now = 1 - step / steps
    t_next = max(1 - (step + 1 + td) / steps, 0)
        
    # Predict analog bits m_0 from m_t.
    _, m_pred = mask_decoder(m_t, h, t_now)
        
    # Estimate m at t_next.
    m_t = ddim_step(m_t, m_pred, t_now, t_next)
    
  # Analog bits to masks.
  masks = bit2int(m_pred > 0)
  return masks
\end{lstlisting}
\end{algorithm}

Our main training objective is the conditional denoising of analog bits~\cite{chen2022analog} that represent noisy panoptic masks. Algorithm~\ref{alg:train} gives the training algorithm (with further details in App.~\ref{app:alg}), the key elements of which are introduced below.

\begin{figure*}[t]
    \centering
    \begin{subfigure}{0.1\textwidth}
      \centering
      $b=1.0$
    \end{subfigure}
    \begin{subfigure}{0.13\textwidth}
      \centering
      \includegraphics[width=\linewidth]{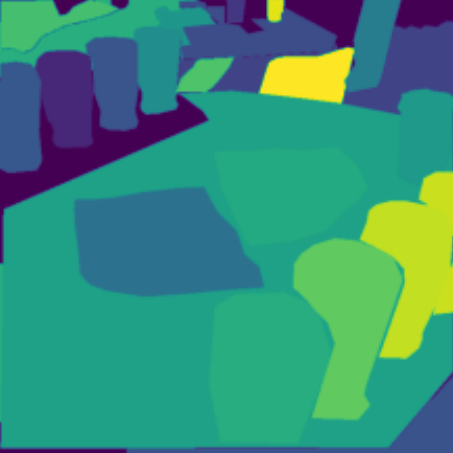}
      \caption{$t=0.1$.}
    \end{subfigure}
    ~~
    \begin{subfigure}{0.13\textwidth}
      \centering
      \includegraphics[width=\linewidth]{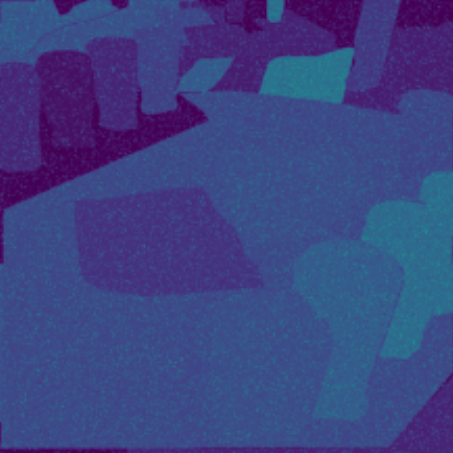}
      \caption{$t=0.3$.}
    \end{subfigure}
    ~~
    \begin{subfigure}{0.13\textwidth}
      \centering
      \includegraphics[width=\linewidth]{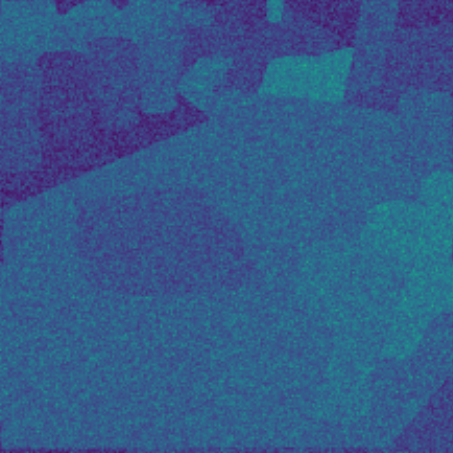}
      \caption{$t=0.5$.}
    \end{subfigure}
    ~~
    \begin{subfigure}{0.13\textwidth}
      \centering
      \includegraphics[width=\linewidth]{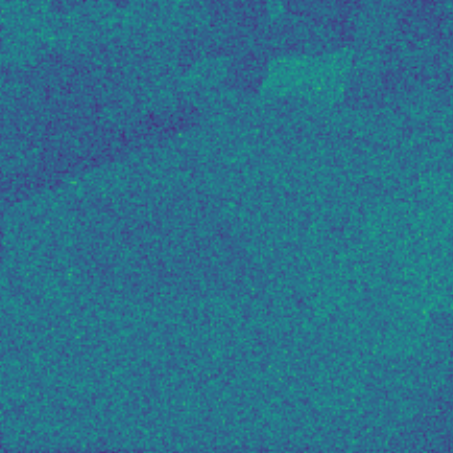}
      \caption{$t=0.7$.}
    \end{subfigure}
    ~~
    \begin{subfigure}{0.13\textwidth}
      \centering
      \includegraphics[width=\linewidth]{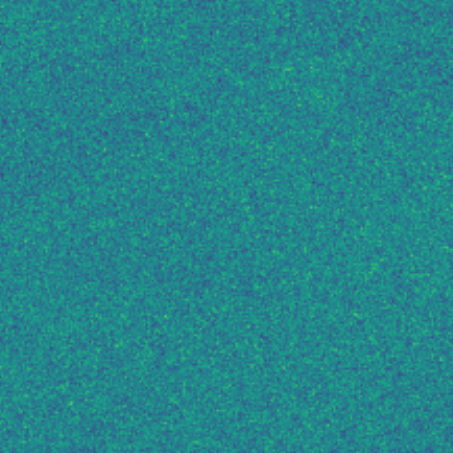}
      \caption{$t=0.9$.}
    \end{subfigure}
    \\
    \begin{subfigure}{0.1\textwidth}
      \centering
      $b=0.1$
    \end{subfigure}
    \begin{subfigure}{0.13\textwidth}
      \centering
      \includegraphics[width=\linewidth]{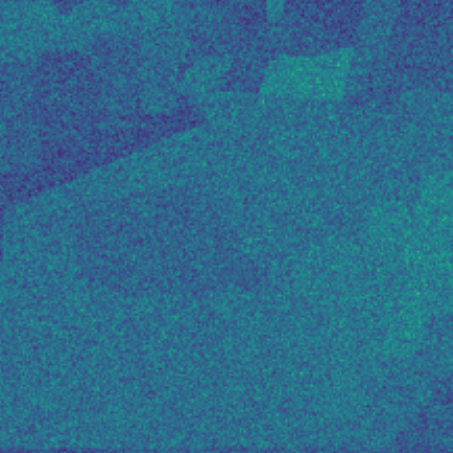}
      \caption{$t=0.1$.}
    \end{subfigure}
    ~~
    \begin{subfigure}{0.13\textwidth}
      \centering
      \includegraphics[width=\linewidth]{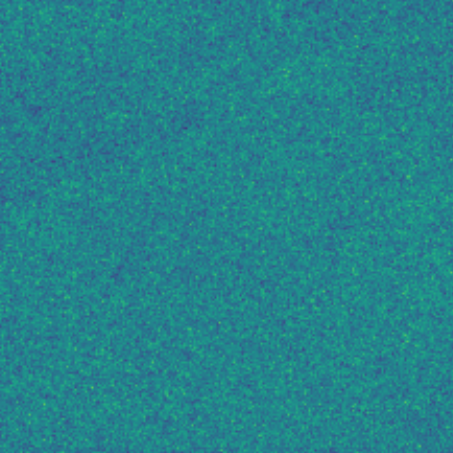}
      \caption{$t=0.3$.}
    \end{subfigure}
    ~~
    \begin{subfigure}{0.13\textwidth}
      \centering
      \includegraphics[width=\linewidth]{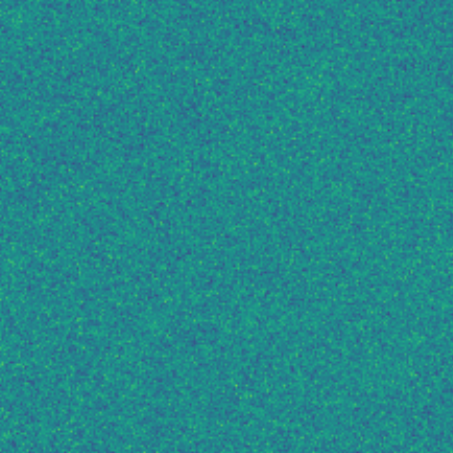}
      \caption{$t=0.5$.}
    \end{subfigure}
    ~~
    \begin{subfigure}{0.13\textwidth}
      \centering
      \includegraphics[width=\linewidth]{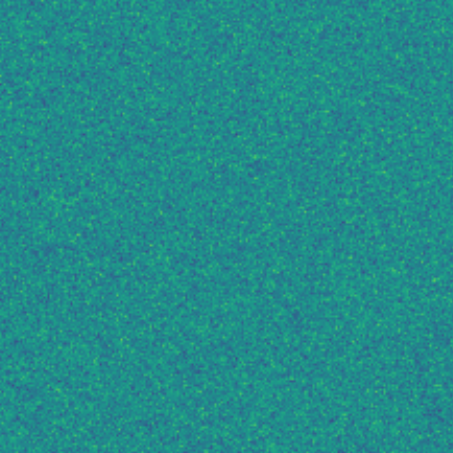}
      \caption{$t=0.7$.}
    \end{subfigure}
    ~~
    \begin{subfigure}{0.13\textwidth}
      \centering
      \includegraphics[width=\linewidth]{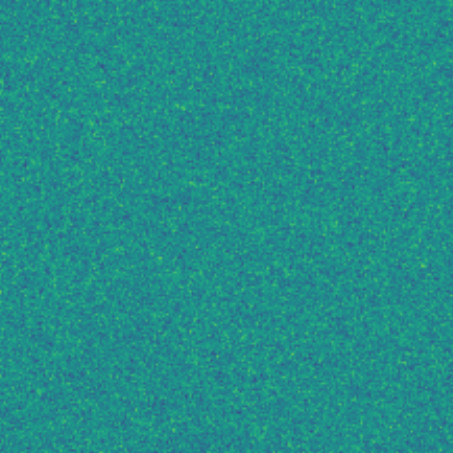}
      \caption{$t=0.9$.}
    \end{subfigure}
    \vspace*{-0.2cm}
    \caption{\label{fig:add_noise}Noisy masks at different time steps under two input scaling factors, $b=1.0$ (top row) and $b=0.1$ (bottom row). Decreasing the input scaling factor leads to smaller signal-to-noise ratio (at the same time step), which gives higher weights to harder cases.}
    \vspace*{-0.2cm}
\end{figure*}

\vspace*{-0.2cm}
\paragraph{Analog Bits with Input Scaling.} The analog bits are real numbers converted from the integers of panoptic masks. When constructing the analog bits, we can shift and scale them into $\{-b, b\}$.  
Typically, $b$ is set to be 1~\cite{chen2022analog} but we find that adjusting this scaling factor has a significant effect on the performance of the model. This scaling factor effectively allows one to adjust  the signal-to-noise ratio of the diffusion process (or the noise schedule), as visualized in Fig.~\ref{fig:add_noise}. With $b=1$, we see that even with a large time step $t=0.7$ (with $t\in[0, 1]$), the signal-to-noise ratio is still relatively high, so the masks are visible to naked eye and the model can easily recover the mask without using the encoded image features.
With $b=0.1$, however, the denoising task becomes significantly harder as the signal-to-noise ratio is reduced. In our study, we find $b=0.1$ works substantially better than the default of 1.0.

\begin{figure*}[t]
    \centering
    \begin{subfigure}{0.2\textwidth}
      \centering
      \includegraphics[width=\linewidth]{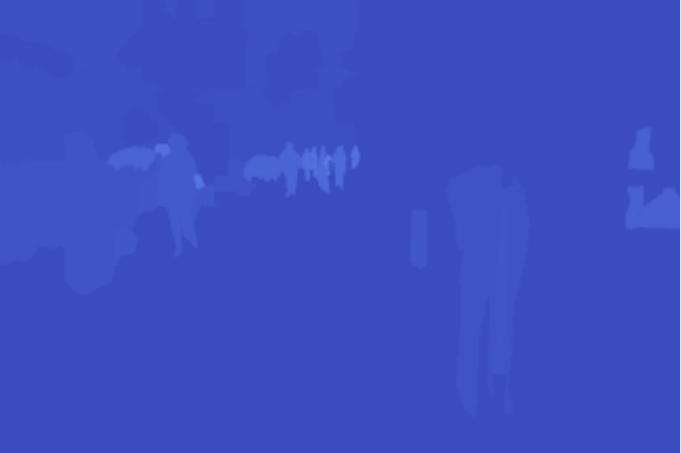}
      \caption{$p=0.1$.}
    \end{subfigure}
    ~~~
    \begin{subfigure}{0.2\textwidth}
      \centering
      \includegraphics[width=\linewidth]{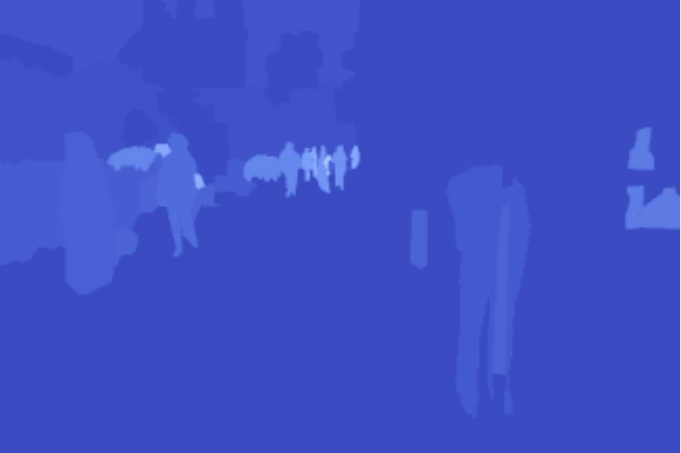}
      \caption{$p=0.2$.}
    \end{subfigure}
    ~~~
    \begin{subfigure}{0.2\textwidth}
      \centering
      \includegraphics[width=\linewidth]{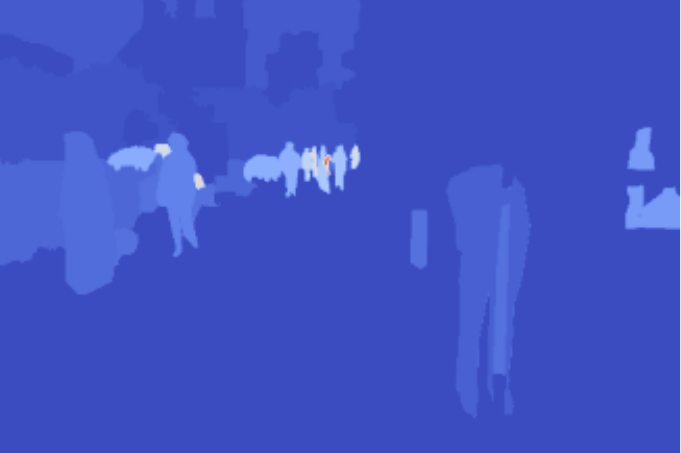}
      \caption{$p=0.3$.}
    \end{subfigure}
    ~~~
    \begin{subfigure}{0.2\textwidth}
      \centering
      \includegraphics[width=\linewidth]{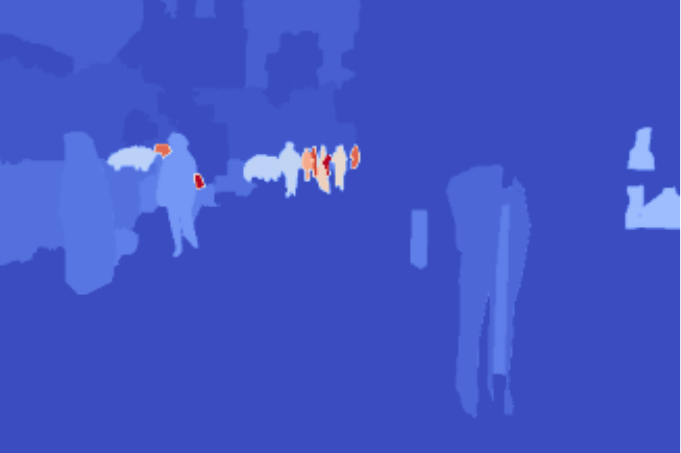}
      \caption{$p=0.4$.}
    \end{subfigure}
    \vspace*{-0.2cm}
    \caption{\label{fig:loss_weighting} The effect of $p$ on loss weighting for panoptic masks. With $p=0$, every mask token is weighted equally (equivalent to no weighting). As $p$ increases, larger weight is given to mask tokens of smaller instances (indicated by warmer colors).}
    \vspace*{-0.2cm}
\end{figure*}

\vspace*{-0.2cm}
\paragraph{Softmax Cross Entropy Loss.} Conventional diffusion models (with or without analog bits) are trained with an $\ell_2$ denoising loss. This works reasonably well for panoptic segmentation, but we also discovered that a loss based on softmax cross entropy yields better performance. 
Although the analog bits are real numbers, they can be seen as a one-hot weighted average of base categories. For example, $\text{`01'}= \alpha_0 \text{`00'} + \alpha_1 \text{`01'} + \alpha_2 \text{`10'} + \alpha_3 \text{`11'}$ where $\alpha_1=1$, and $\alpha_0=\alpha_2=\alpha_3=0$. 
Instead of modeling the analog bits in `01' as real numbers, with a cross entropy loss, the network can directly model the underlying distribution over the four base categories, and use the weighted average to obtain the analog bits. As such, the mask decoder output not only analog bits (\verb|m_pred|), but also the corresponding logits (\verb|m_logits|), $ \tilde{\bm y} \in \mathbb{R}^{H\times W\times K}$, with a cross entropy loss for training; i.e.,
$$
\mathcal{L} = \sum_{i,j,k}\bm y_{ijk} \log\text{softmax}(\tilde{\bm y}_{ijk})
$$
Here, $\bm y$ is the one-hot vector corresponding to class or instance label. 
During inference, we still use analog bits from the mask decoder instead of underlying logits for the reverse diffusion process.

\vspace*{-0.1cm}
\paragraph{Loss Weighting.} Standard generative models for discrete data assign equal weight to all tokens.
For panoptic segmentation, with a loss defined over pixels, this means that large objects will have more influence than small objects.
And this makes learning to segment small instances inefficient.
To mitigate this, we use an adaptive loss to improve the segmentation of small instances by giving higher weights to mask tokens associated with small objects. The specific per-token loss weighting is as follows:
$$
w_{ij} = 1/c_{ij}^p ~, ~\text{and }\,
w'_{ij} =  H * W * w_{ij} / \sum_{ij} w_{ij} ~,
$$
where $c_{ij}$ is the pixel count for the instance at pixel location $(i,j)$, and $p$ is a tunable parameter; uniform weighting occurs when $p=0$, and for $p>0$, a higher weight is assigned to mask tokens of small instances.
Fig.~\ref{fig:loss_weighting} demonstrate the effects of $p$ in weighting different mask tokens.

\subsection{Inference Algorithm}

Algorithm~\ref{alg:sample} summarizes the inference procedure.
Given an input image, the model starts with random noise as the initial analog bits, and gradually refines its estimates to be closer to that of good panoptic masks. Like Bit Diffusion~\cite{chen2022analog}, we use asymmetric time intervals (controlled by a single parameter \verb|td|) that is adjustable during inference time. It is worth noting that the encoder is only run once, so the cost of multiple iterations depends on the decoder alone.

\subsection{Extension to Videos}

\begin{figure}[t]
    \centering
    \includegraphics[width=.49\textwidth]{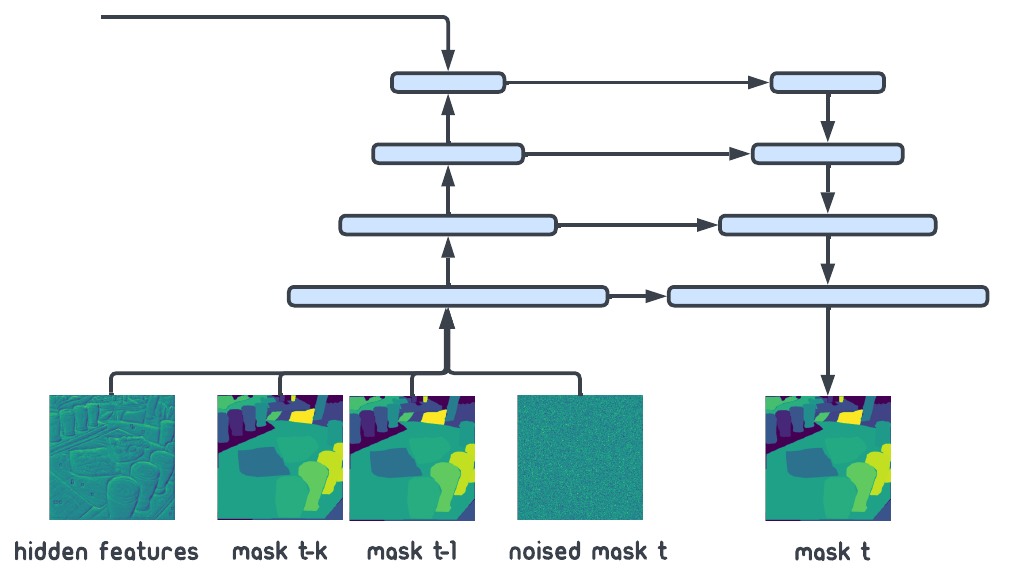}
    \caption{Mask decoder extended for video settings. The image conditional signal to the mask decoder is concatenated with mask predictions from previous frames of the video.}
    \vspace*{-0.2cm}
    \label{fig:architecture_video}
    \vspace*{-0.2cm}
\end{figure}

Our image-conditional panoptic mask modeling with $p(\bm m|\bm x)$ is directly applicable for video panoptic segmentation by considering 3D masks (with an extra time dimension) given a video.
To adapt for online/streaming video settings, 
we can instead model $p(\bm m_t|\bm x_t, \bm m_{t-1}, \bm m_{t-k})$, thereby generating panoptic masks conditioned on the image and past mask predictions. This change can be easily implemented by concatenating the past panoptic masks ($\bm m_{t-1}, \bm m_{t-k}$) with existing noisy masks, as demonstrated in Fig.~\ref{fig:architecture_video}; $k$ is a hyper-parameters of the model.
Other than this minor change, the model remains the same as that above, which is simple and allows one to fine-tune an image panoptic model for video.

With the past-conditional generation (using the denoising objective), the model automatically learns to track and segment instances across frames, without requiring explicit instance matching through time.
Having an iterative refinement procedure also makes our framework convenient to adapt in a streaming video setting where there are strong dependencies across adjacent frames. We expect fewer inference steps are needed for good segmentation when there are relatively small changes in video frames, so it may be possible to set the refinement steps adaptively.

 \section{Experiments}

\begin{table*}[t!]
    \centering
    \small
    \begin{tabular}{r|c|c|c|c|c}
        \toprule
        Method & Backbone &\# of Params & PQ & PQ$^\text{thing}$ & PQ$^\text{stuff}$ \\
        \midrule
        \multicolumn{6}{l}{\textit{Specialist approaches:\hfill}} \\
        MaskFormer~\cite{cheng2021per} & ResNet-50 & 45M & 46.5 & 51.0 & 39.8 \\
        K-Net~\cite{zhang2021k} & ResNet-50 & - & 47.1 & 51.7 & 40.3 \\
        CMT-DeepLab~\cite{yu2022cmt} & ResNet-50 & - & 48.5 & - & - \\
        Panoptic SegFormer~\cite{li2022panoptic} & ResNet-50 & 51M & 49.6 & 54.4 & 42.4 \\
        Mask2Former~\cite{cheng2021masked} & ResNet-50 & 44M & 51.9 & 57.7 & 43.0 \\
        kMaX-DeepLab~\cite{yu2022kmeans} & ResNet-50 & 57M & 53.0 & 58.3 & 44.9 \\
        DETR~\cite{carion2020end}       & ResNet-101 & 61.8M & 45.1 & 50.5 & 37.0 \\
Mask2Former~\cite{cheng2021mask2former} & Swin-L & 216M & 57.8 & - & - \\ 
kMaX-DeepLab~\cite{yu2022kmeans} & ConvNeXt-L & 232M & 58.1 & 64.3 & 48.8 \\
        MasK DINO~\cite{li2022mask} & Swin-L & 223M & 59.4 & - & - \\
        \midrule
        \multicolumn{6}{l}{\textit{Generalist approaches:\hfill}} \\
        ~~UViM~\cite{UViM} & ViT & 939M & 45.8 & - & - \\
        ~~\ptsd (steps=5) & ResNet-50  &  94.5M & 47.5 & 52.2 & 40.3 \\
        ~~\ptsd (steps=10) & ResNet-50 &  94.5M & 49.4 & 54.4 & 41.9\\
        ~~\ptsd (steps=20) & ResNet-50  &  94.5M & 50.3 & 55.3 & 42.9\\
        ~~\ptsd (steps=50) & ResNet-50  &  94.5M & 50.2 & 55.1 & 42.8\\
        \bottomrule
    \end{tabular}
    \vspace*{-0.2cm}
    \caption{Results on MS-COCO. \ptsd achieves competitive results to state-of-the-art specialist models with ResNet-50 backbone.}
\vspace*{-0.2cm}
    \label{tab:main_result}
\end{table*}

\begin{table*}[]
    \centering
    \small
    \begin{tabular}{c|c|ccccc}
         \toprule
         Method & Backbone & $\mathcal{J\&F}$ & $\mathcal{J}$-Mean & $\mathcal{J}$-Recall & $\mathcal{F}$-mean & $\mathcal{F}$-Recall \\
         \midrule
        \multicolumn{7}{l}{\textit{Specialist approaches:\hfill}} \\
         RVOS \cite{ventura2019rvos}            & ResNet-101    & 41.2	& 36.8	& 40.2	& 45.7	& 46.4 \\
STEm-Seg \cite{athar2020stem}          & ResNet-101    & 64.7  & 61.5  & 70.4  & 67.8  & 75.5 \\
         MAST \cite{Lai_2020_CVPR}              & ResNet-18     & 65.5  & 63.3  & 73.2  & 67.6  & 77.7 \\
         UnOVOST \cite{Luiten_2020_WACV}        & ResNet-101    & 67.9  & 66.4  & 76.4  & 69.3  & 76.9 \\
         Propose-Reduce \cite{Lin_2021_ICCV}    & ResNeXt-101   & 70.4  & 67.0  & -     & 73.8  & -    \\
         \midrule
        \multicolumn{7}{l}{\textit{Generalist approaches:\hfill}} \\
         \ptsd (ours)                           & ResNet-50     & 68.4  & 65.1  & 70.6  & 71.7  & 77.1 \\
         \bottomrule
    \end{tabular}
    \vspace*{-0.2cm}
    \caption{Results of unsupervised video object segmentation on DAVIS 2017 validation set.}
\label{tab:video_main}
\end{table*}

\subsection{Setup and Implementation Details}
\label{sub:exp_setup}

\vspace*{-0.1cm}
\paragraph{Datasets.} For image panoptic segmentation, we conduct experiments on the two commonly used benchmarks: MS-COCO~\cite{lin2014microsoft} and Cityscapes~\cite{cordts2016cityscapes}. MS-COCO contains approximately 118K training images and 5K validation images used for evaluation. Cityscapes contains 2975 images for training, 500 for validation and 1525 for testing. We report results on Cityscapes \textit{val} set, following most existing papers.
For expedience we conduct most model ablations on MS-COCO. 
Finally, for video segmentation we use DAVIS~\cite{perazzi2016benchmark} in the challenging unsupervised setting, with no segmentation provided at test time.
DAVIS comprises 60 training videos and 30 validation videos for evaluation.

\vspace*{-0.2cm}
\paragraph{Training.} MS-COCO is larger and more diverse than Cityscapes and DAVIS. Thus we mainly train on MS-COCO, and then transfer trained models to Cityscapes and DAVIS with fine-tuning (at a single resolution).
We first separately pre-train the image encoder and mask decoder before training the image-conditional mask generation on MS-COCO. The image encoder is taken from the Pix2Seq~\cite{chen2021pix2seq} object detection checkpoint, pre-trained on objects365~\cite{shao2019objects365}.
It comprises a ResNet-50~\cite{he2016deep} backbone, and 6-layer 512-dim Transformer~\cite{vaswani2017attention} encoder layers. 
We also augment the image encoder with a few convolutional upsampling layers to increase its resolution and incorporate features at different layers.
The mask decoder is a TransUNet~\cite{chen2021transunet} with base dimension 128, and channel multipliers of 1$\times$,1$\times$,2$\times$,2$\times$, followed by 6-layer 512-dim Transformer~\cite{vaswani2017attention} decoder layers. It is pre-trained on MS-COCO as an unconditional mask generation model without images.

Directly training our model on high resolution images and panoptic masks can be expensive as the existing architecture scales quadratically with resolution. So on MS-COCO, we train the model with increasing resolutions, similar to~\cite{tan2021efficientnetv2,touvron2019fixing,training2018}.
We first train at a lower resolution (256$\times$256 for images; 128$\times$128 for masks) for 800 epochs with a batch size of 512 and scale jittering~\cite{ghiasi2021simple,wu2019detectron2} of strength $[1.0, 3.0]$. We then continue train the model at full resolution (1024$\times$1024 for images; 512$\times$512 for masks) for only 15 epochs with a batch size of 16 without augmentation.
This works well, as both convolution networks and transformers with sin-cos positional encoding generalize well across resolutions.
More details on hyper-parameter settings for training can be found in Appendix~\ref{app:hparams}.

\vspace*{-0.2cm}
\paragraph{Inference.} We use DDIM updating rules~\cite{song2020denoising} for sampling.  By default we use 20 sampling steps for MS-COCO. We find that setting \verb|td|$=2.0$ yields near optimal results. We discard instance predictions with fewer than 80 pixels.
For inference on DAVIS, we use 32 sampling steps for the first frame and 8 steps for subsequent frames. We set \verb|td|$=1$ and discard instance predictions with fewer than 10 pixels.

\subsection{Main Results}

We compare with two families of state-of-the-art methods, i.e., specialist approaches and generalist approaches.
Table~\ref{tab:main_result} summarizes results for MS-COCO.
\ptsd achieves competitive Panoptic Quality (PQ) to state-of-the-art methods with the ResNet-50 backbone. When compared with other recent generalist models such as UViM~\cite{UViM}, our model performs significantly better while being much more efficient. Similar results are obtained for Cityscape, the details of which are given in Appendix~\ref{app:results_cityscape}.

Table~\ref{tab:video_main} compares \ptsd to state-of-the-art methods on unsupervised video object segmentation on DAVIS, using the standard $\mathcal{J\&F}$ metrics~\cite{perazzi2016benchmark}.
Baselines do not include other generalist models as they are not directly applicable to the task. Our method achieves results on par with state-of-the-art methods without specialized designs. 

\subsection{Ablations on Training}

Ablations on model training are performed with MS-COCO dataset. 
To reduce the computation cost while still being able to demonstrate the performance differences in design choices, 
we train the model for 100 epochs with a batch size of 128 in a single-resolution stage (512$\times$512 image size, and 256$\times$256 mask size).

\textbf{Scaling of Analog Bits.} 
Table~\ref{tab:b_scale} shows the dependence of PQ results on input scaling of analog bits.
The scale factor of 0.1 used here yield results that outperform the  standard scaling of 1.0 in previous work~\cite{chen2022analog}.

\begin{table}[t]
    \centering
    \small
    \vspace*{-0.1cm}
    \begin{tabular}{c|cccc}
         \toprule
         Input scaling & 0.03 & 0.1 & 0.3 & 1.0 \\
         \midrule
         PQ & 40.8 & 43.9 & 38.7 & 21.3\\
         \bottomrule
    \end{tabular}
    \vspace*{-0.2cm}
    \caption{Ablation on input scaling}
    \vspace*{-0.2cm}
    \label{tab:b_scale}
\end{table}

\textbf{Loss Functions.}
Table~\ref{tab:loss_f} compares our proposed cross entropy loss to the $\ell_2$ loss normally used by diffusion models. Interestingly, the cross entropy loss yields substantial gains over $\ell_2$.

\begin{table}[t]
    \centering
    \small
    \vspace*{-0.1cm}
    \begin{tabular}{c|cc}
         \toprule
         Loss function & $\ell_2$ Regression & Cross Entropy \\
         \midrule
         PQ & 41.9 & 43.9 \\
         \bottomrule
    \end{tabular}
    \vspace*{-0.2cm}
    \caption{Ablation on loss function.}
    \vspace*{-0.2cm}
    \label{tab:loss_f}
\end{table}

\textbf{Loss weighting.}
Table~\ref{tab:loss_w} shows the effects of $p$ for loss weighting. Weighting with $p=0.2$ appears near optimal and clearly outperforms uniform weighting ($p=0$).

\begin{table}[t]
    \centering
    \small
    \vspace*{-0.1cm}
    \begin{tabular}{c|cccc}
         \toprule
         Loss weight $p$ & 0 & 0.2 & 0.4 & 0.6\\
         \midrule
PQ & 40.4 & 43.9 & 43.7 & 41.3\\  \bottomrule
    \end{tabular}
    \vspace*{-0.2cm}
    \caption{Ablation on loss weighting.}
    \vspace*{-0.2cm}
    \label{tab:loss_w}
\end{table}

\vspace*{-0.2cm}
\subsection{Ablations on Inference}

\begin{figure}[t]
    \centering
    \begin{subfigure}{0.32\linewidth}
      \centering
      \includegraphics[width=\linewidth]{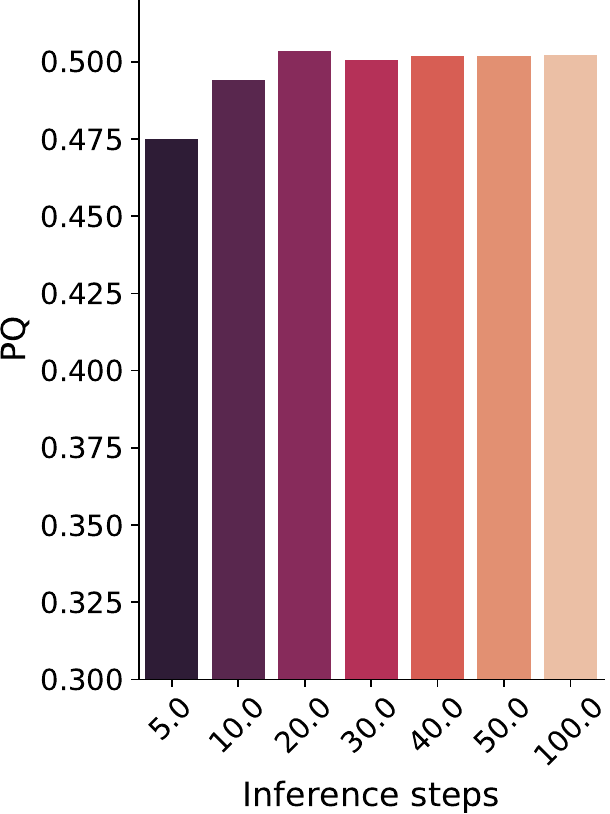}
      \caption{\label{fig:infer_ablations1}}
    \end{subfigure}
    \begin{subfigure}{0.32\linewidth}
      \centering
      \includegraphics[width=\linewidth]{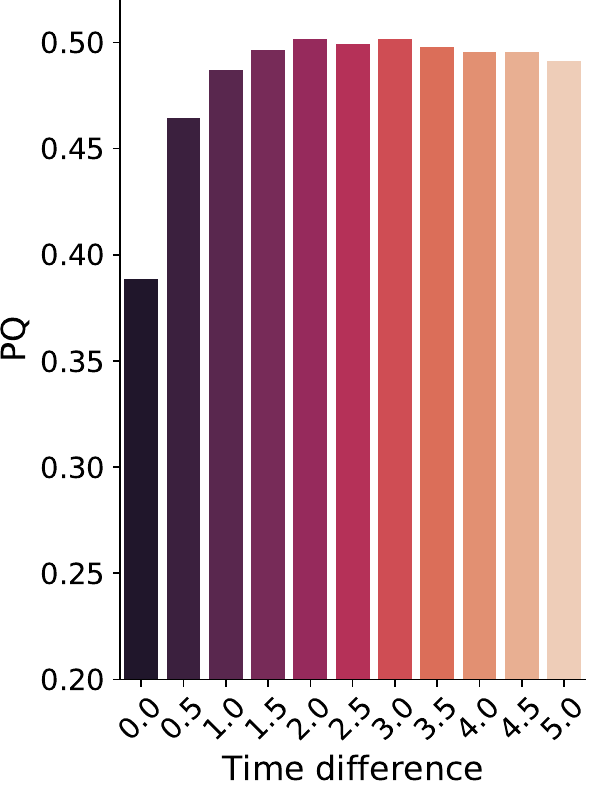}
      \caption{\label{fig:infer_ablations2}}
    \end{subfigure}
    \begin{subfigure}{0.32\linewidth}
      \centering
      \includegraphics[width=\linewidth]{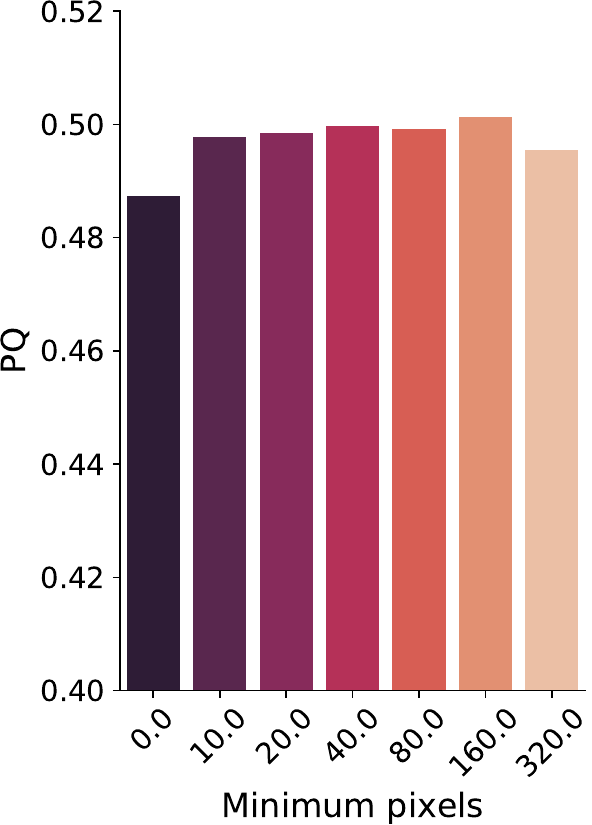}
      \caption{\label{fig:infer_ablations3}}
    \end{subfigure}
    \vspace*{-0.2cm}
    \caption{\label{fig:infer_ablations} Inference ablations on MS-COCO.}
    \vspace*{-0.2cm}
\end{figure}

Figure ~\ref{fig:infer_ablations} explores  the dependence of model performance on hyper-parameter choices during inference (sampling), namely, the number of inference steps, time differences and threshold on the minimum size of instance regions, all on MS-COCO. 
Specifically, Fig.~\ref{fig:infer_ablations1} shows that inference steps of 20 seems sufficient for near optimal performance on MS-COCO.
Fig.~\ref{fig:infer_ablations2} shows that the \verb|td| parameter as in asymmetric time intervals~\cite{chen2022analog} has a significant impact, with intermediate values (e.g., 2-3) yielding the best results. 
Fig.~\ref{fig:infer_ablations3} shows that the right choice of threshold on small instances leads to small performance gains.

\begin{figure}[t]
    \centering
    \includegraphics[width=0.8\linewidth]{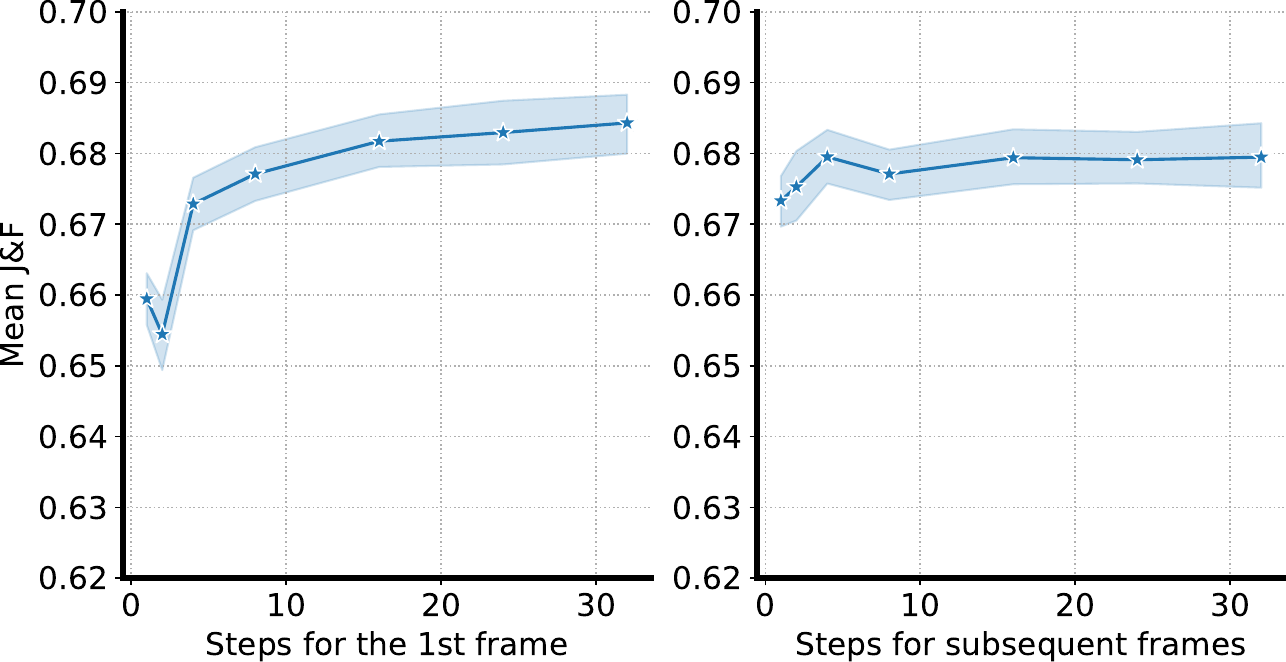}
    \caption{Effect of inference steps on DAVIS. Left: we vary the number of steps for the 1st frame while using a fixed of 8 steps for the remaining frames. Right: we use 8 steps for the 1st frame, while varying the number of steps for the remaining frames. The first frame requires more inference steps due to the cold start.}
    \vspace*{-0.2cm}
    \label{fig:davis_iterations}
    \vspace*{-0.2cm}
\end{figure}

Figure~\ref{fig:davis_iterations} shows how performance varies with the number of inference steps for the first frame, and for the remaining frames, on DAVIS video dataset. We find that more inference steps are helpful for the first frame compared to subsequent frames of video. Therefore, we can reduce the total number of steps by using more steps for the first frames (e.g., 32), and fewer steps for subsequent frames (e.g., 8). It is also worth noting that even using 8 steps for the first frame and only 1 step for each subsequent frame, the model still achieves an impressive $\mathcal{J\&F}$ of 67.3.

\subsection{Case study}

\begin{figure*}[t]
    \centering
    \begin{subfigure}{0.3\textwidth}
      \centering
      Image
    \end{subfigure}
    \begin{subfigure}{0.3\textwidth}
      \centering
      Prediction
    \end{subfigure}
    \begin{subfigure}{0.3\textwidth}
      \centering
      Groundtruth
    \end{subfigure}
\begin{subfigure}{0.29\textwidth}
      \centering
      \includegraphics[width=\linewidth]{figures/case_000000018380_img.jpeg}
    \end{subfigure}
    \begin{subfigure}{0.29\textwidth}
      \centering
      \includegraphics[width=\linewidth]{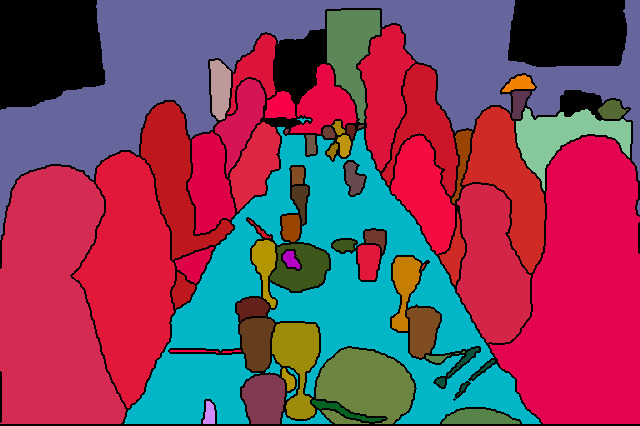}
    \end{subfigure}
    \begin{subfigure}{0.29\textwidth}
      \centering
      \includegraphics[width=\linewidth]{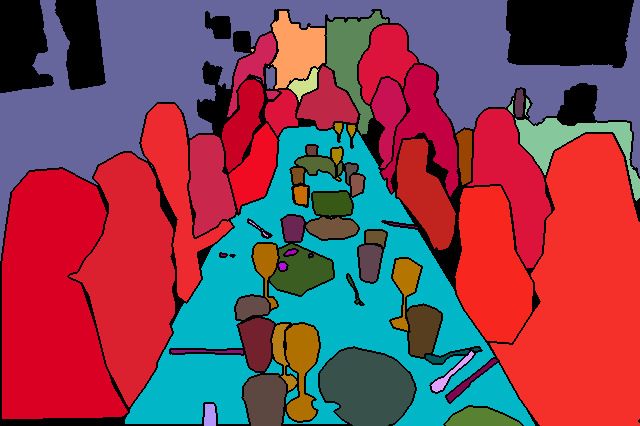}
    \end{subfigure}
    \vspace*{-0.1cm}
    \caption{\label{fig:main_case_coco} Predictions on MS-COCO \textit{val} set.}
    \vspace*{-0.1cm}
\end{figure*}

\begin{figure*}[t!]
    \centering
    \begin{subfigure}{0.32\textwidth}
      \centering
      Image
    \end{subfigure}
    \begin{subfigure}{0.32\textwidth}
      \centering
      Prediction
    \end{subfigure}
    \begin{subfigure}{0.32\textwidth}
      \centering
      Groundtruth
    \end{subfigure}

    \begin{subfigure}{0.32\textwidth}
      \centering
      \includegraphics[width=\linewidth]{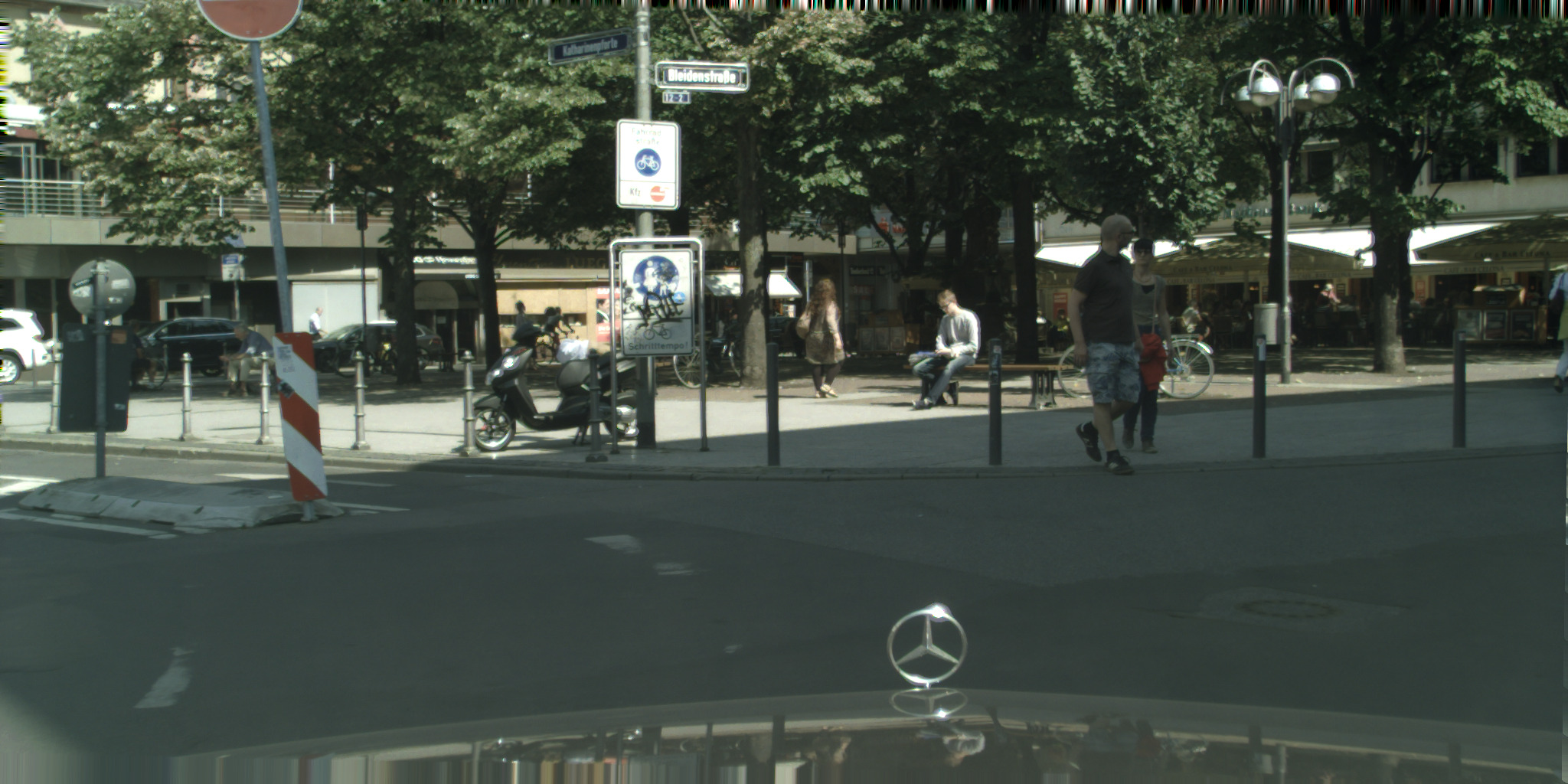}
    \end{subfigure}
    \begin{subfigure}{0.32\textwidth}
      \centering
      \includegraphics[width=\linewidth]{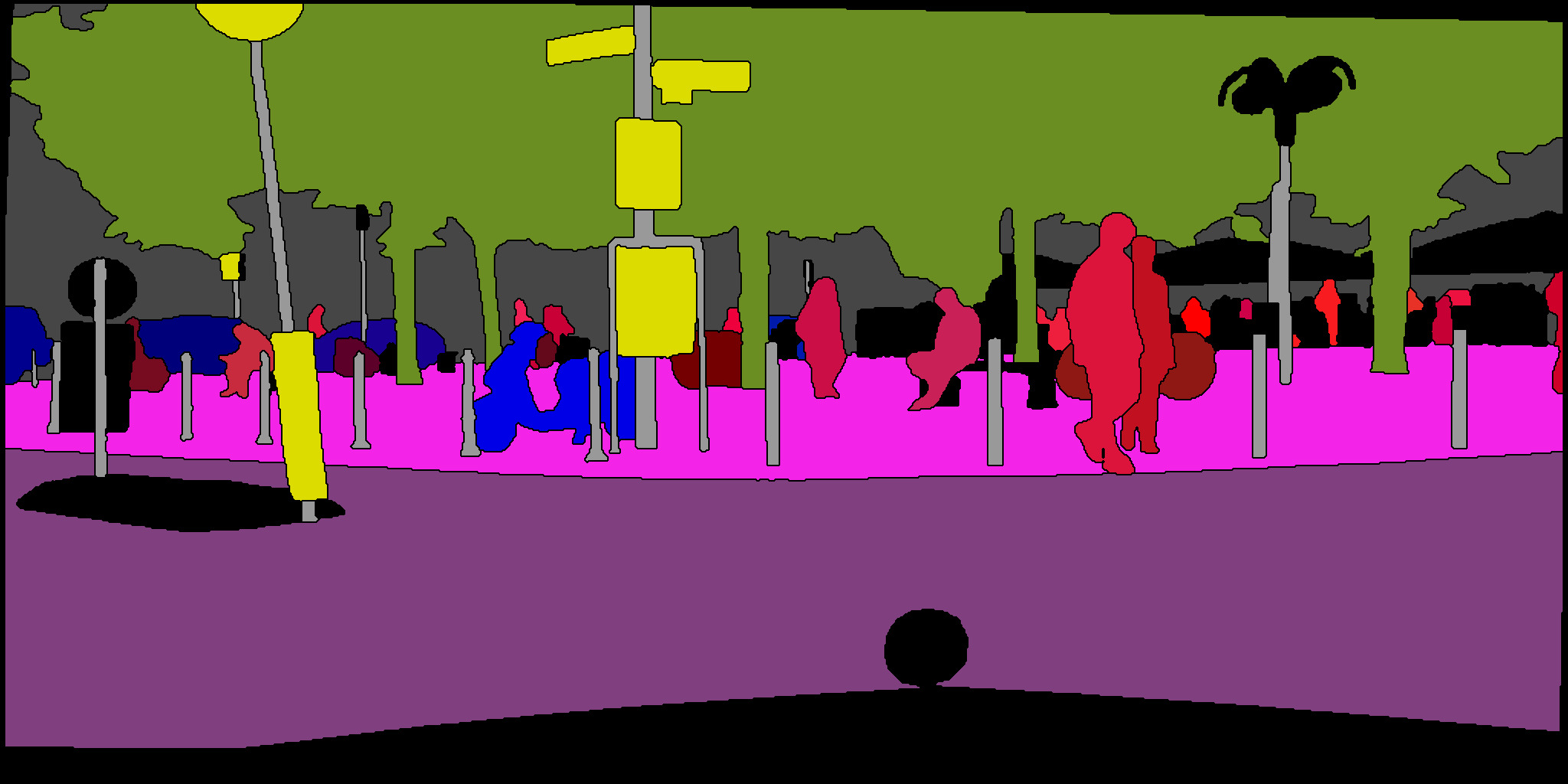}
    \end{subfigure}
    \begin{subfigure}{0.32\textwidth}
      \centering
      \includegraphics[width=\linewidth]{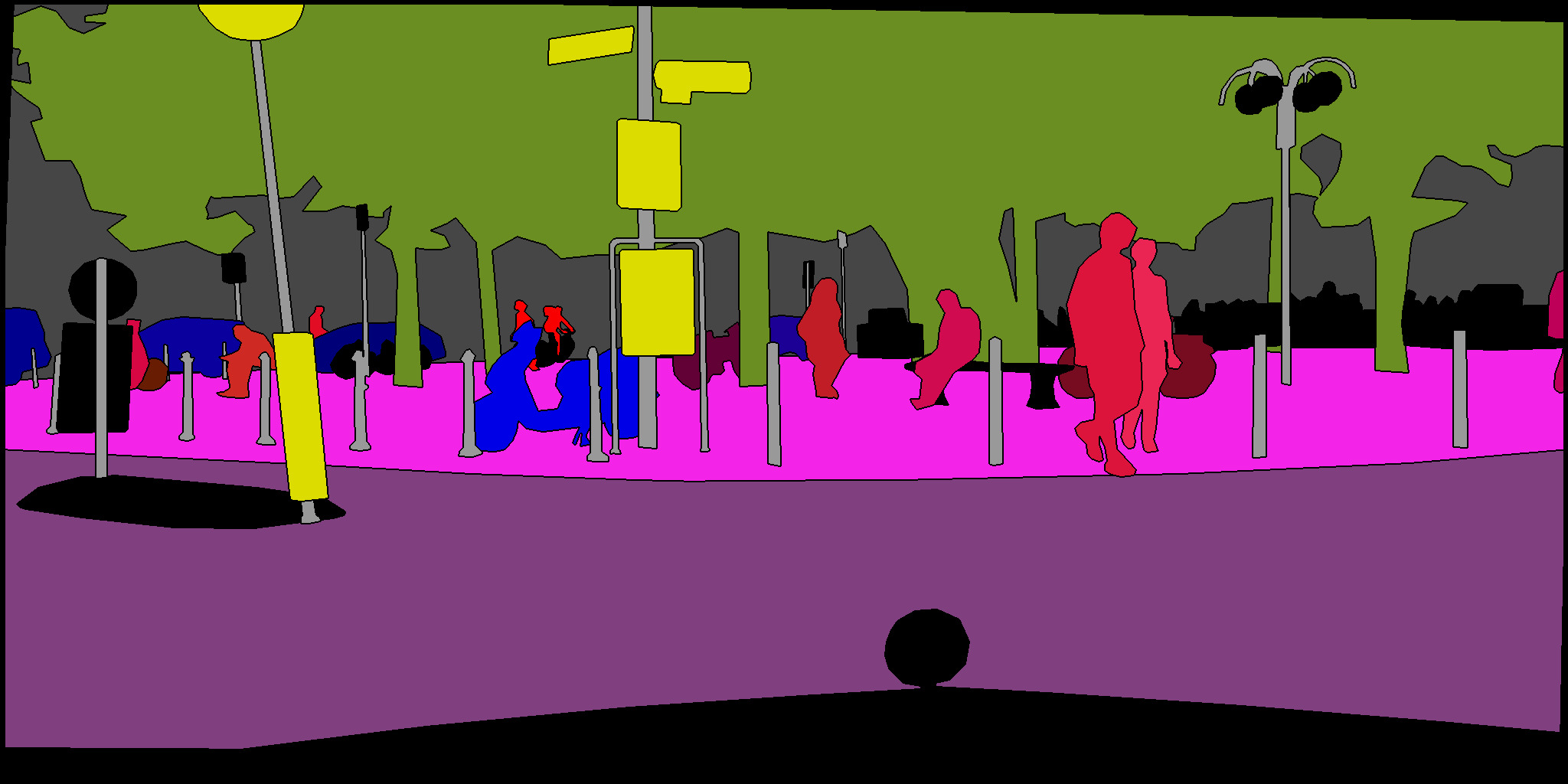}
    \end{subfigure}

    \begin{subfigure}{0.32\textwidth}
      \centering
      \includegraphics[width=\linewidth]{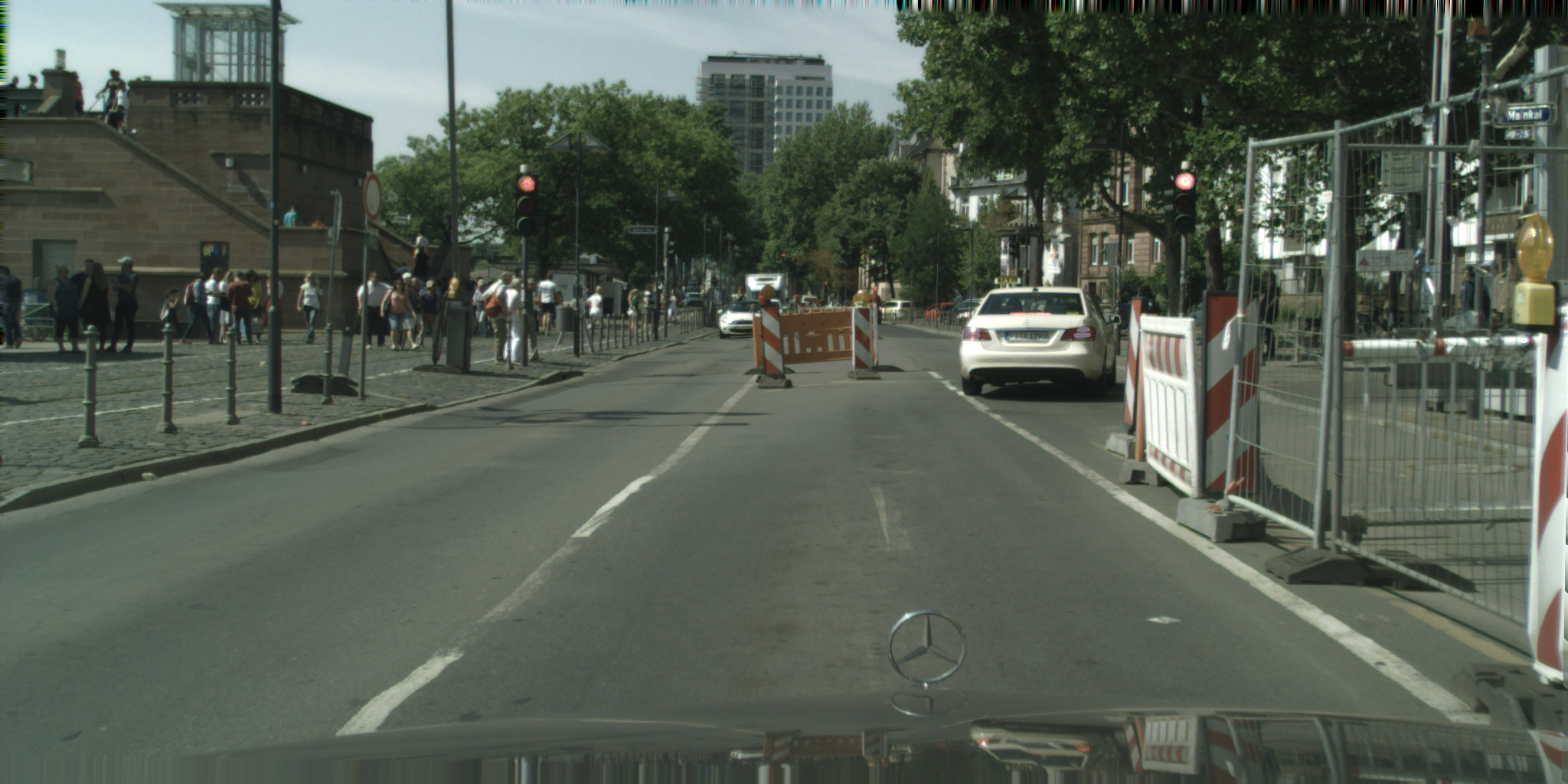}
    \end{subfigure}
    \begin{subfigure}{0.32\textwidth}
      \centering
      \includegraphics[width=\linewidth]{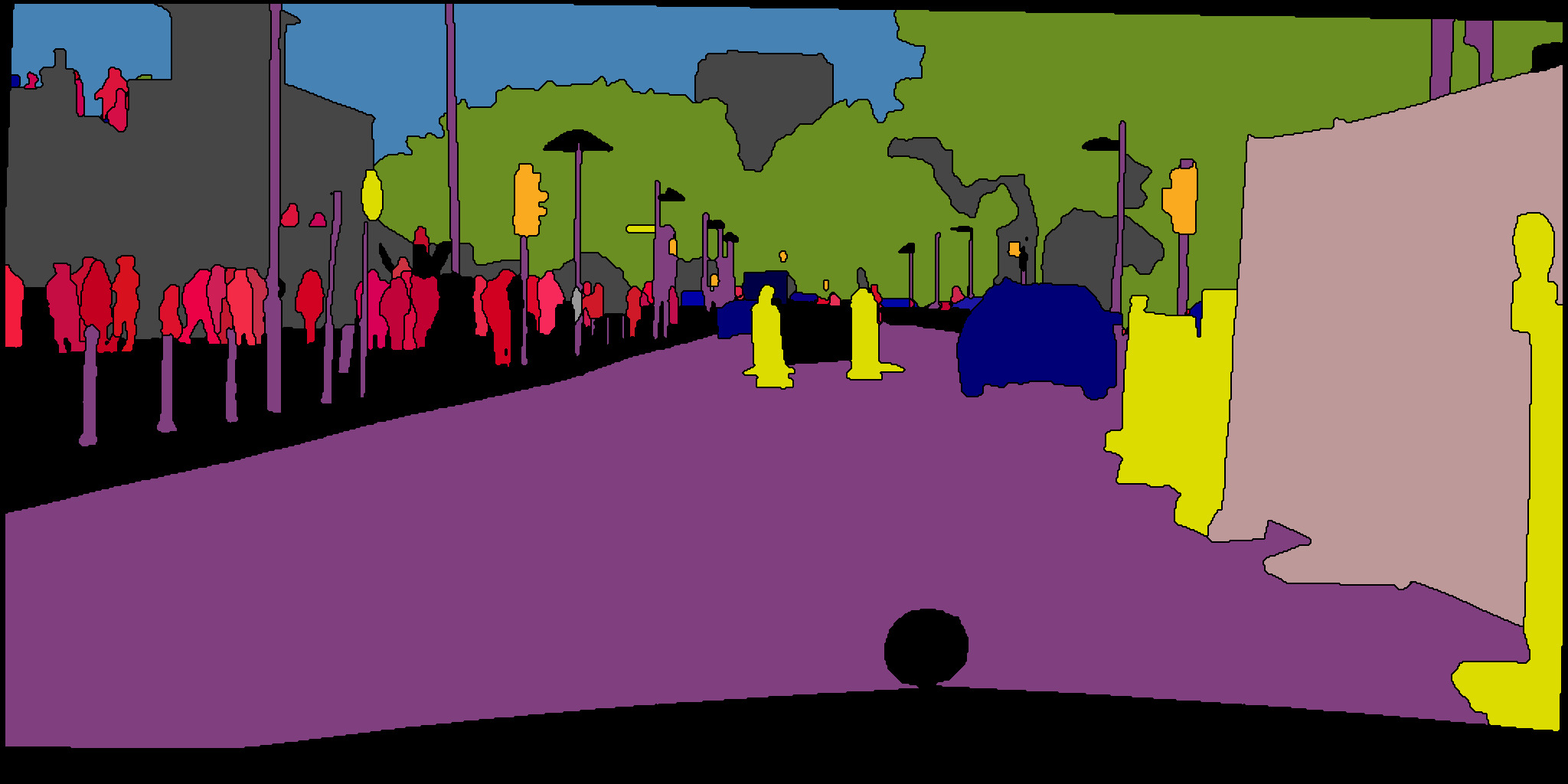}
    \end{subfigure}
    \begin{subfigure}{0.32\textwidth}
      \centering
      \includegraphics[width=\linewidth]{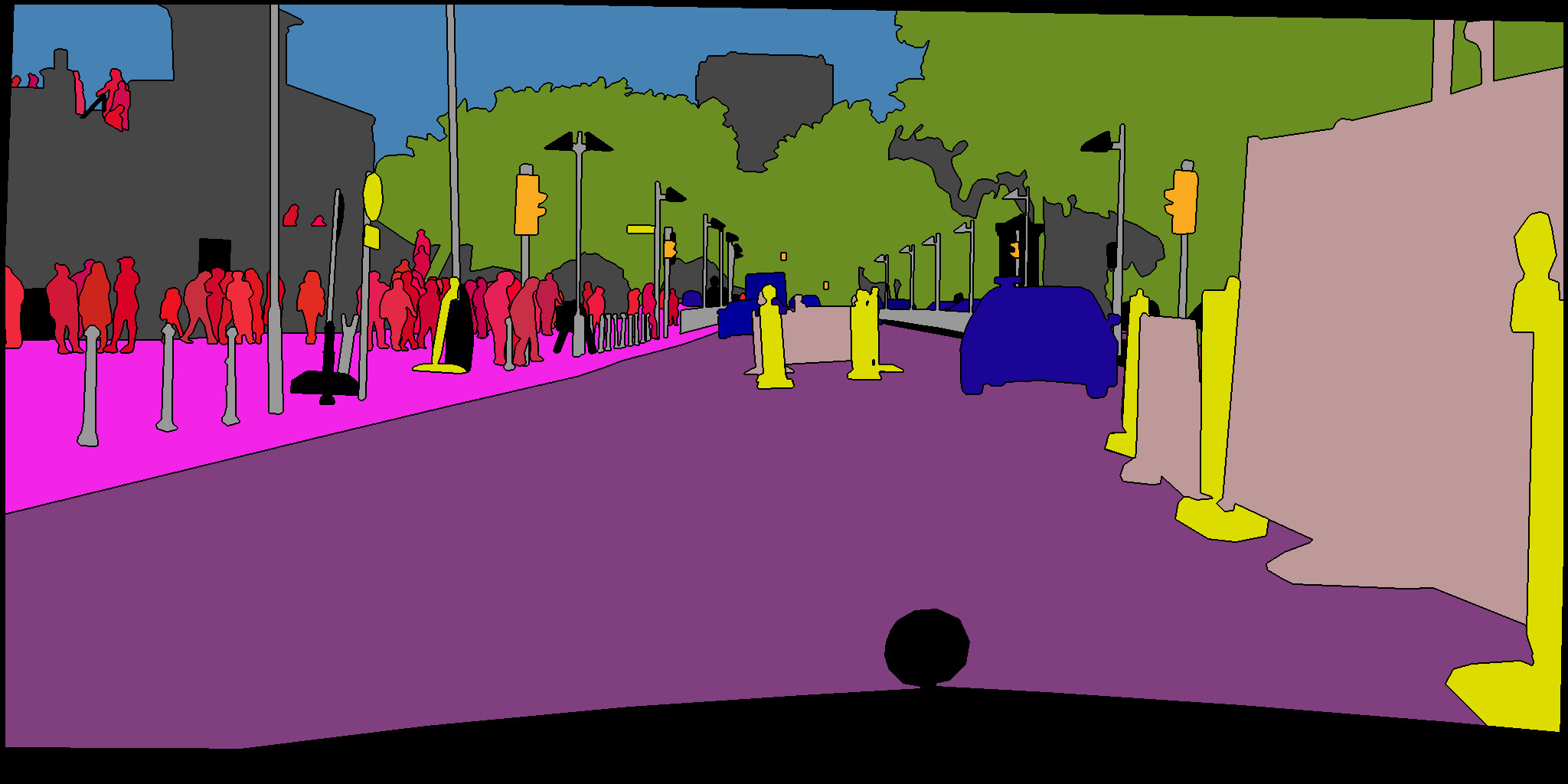}
    \end{subfigure}

    \vspace*{-0.1cm}
    \caption{\label{fig:main_case_city} Predictions on Cityscapes \textit{val} set.}
    \vspace*{-0.1cm}
\end{figure*}

\begin{figure*}[t!]
    \centering
    \begin{subfigure}{0.24\textwidth}
      \centering
      \includegraphics[width=\linewidth]{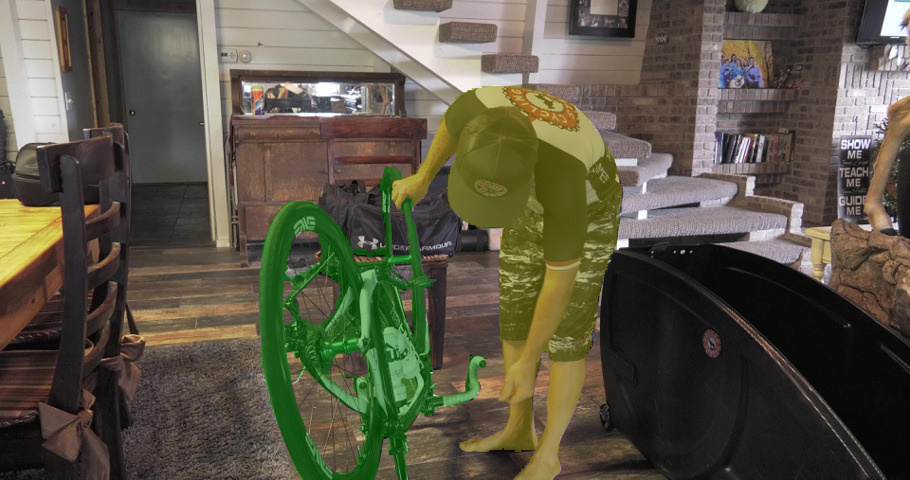}
    \end{subfigure}
    \begin{subfigure}{0.24\textwidth}
      \centering
      \includegraphics[width=\linewidth]{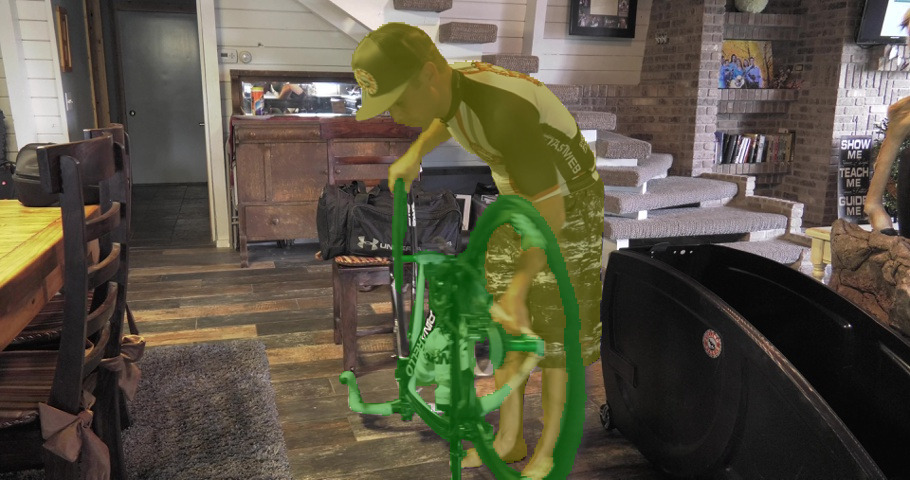}
    \end{subfigure}
    \begin{subfigure}{0.24\textwidth}
      \centering
      \includegraphics[width=\linewidth]{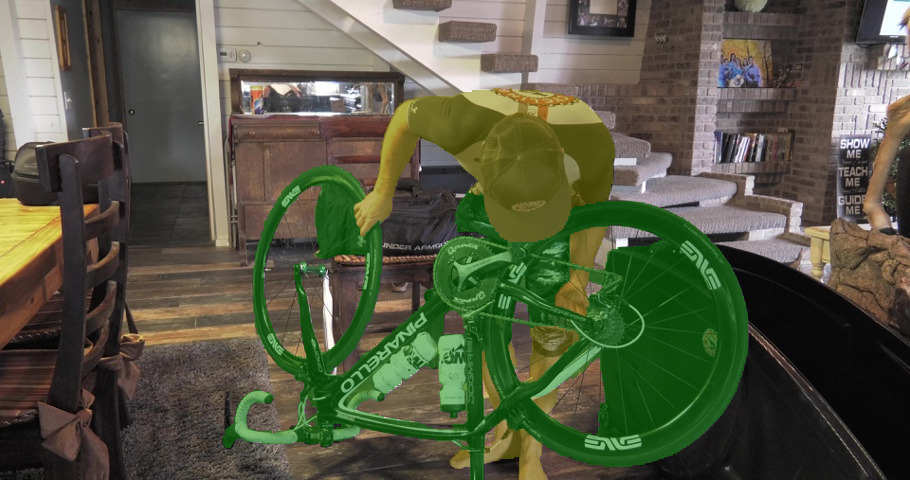}
    \end{subfigure}
    \begin{subfigure}{0.24\textwidth}
      \centering
      \includegraphics[width=\linewidth]{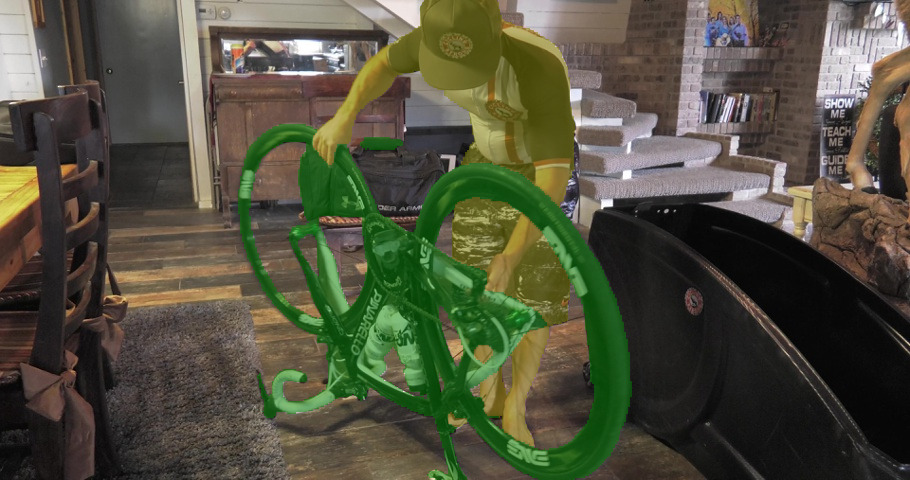}
    \end{subfigure}
    \begin{subfigure}{0.24\textwidth}
      \centering
      \includegraphics[width=\linewidth]{figures/soapbox_0.jpg}
    \end{subfigure}
    \begin{subfigure}{0.24\textwidth}
      \centering
      \includegraphics[width=\linewidth]{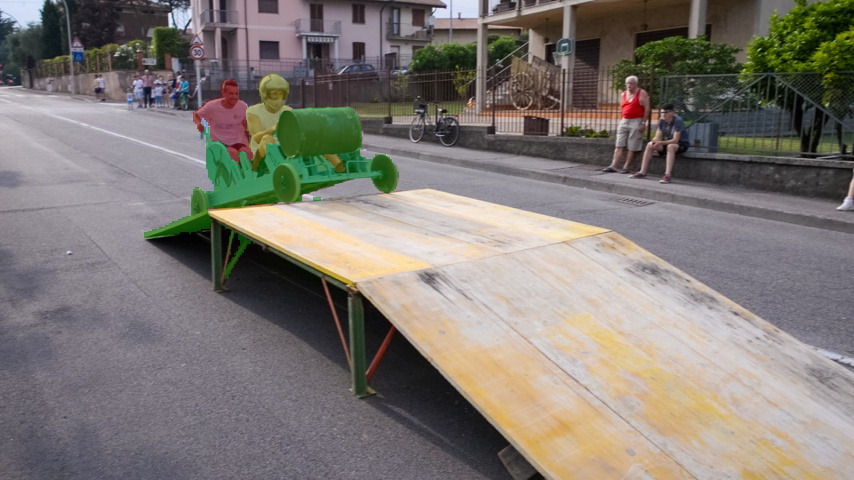}
    \end{subfigure}
    \begin{subfigure}{0.24\textwidth}
      \centering
      \includegraphics[width=\linewidth]{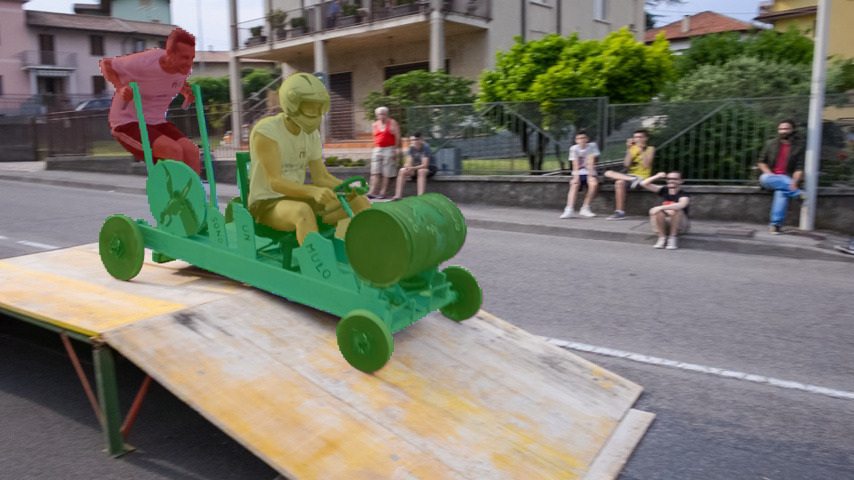}
    \end{subfigure}
    \begin{subfigure}{0.24\textwidth}
      \centering
      \includegraphics[width=\linewidth]{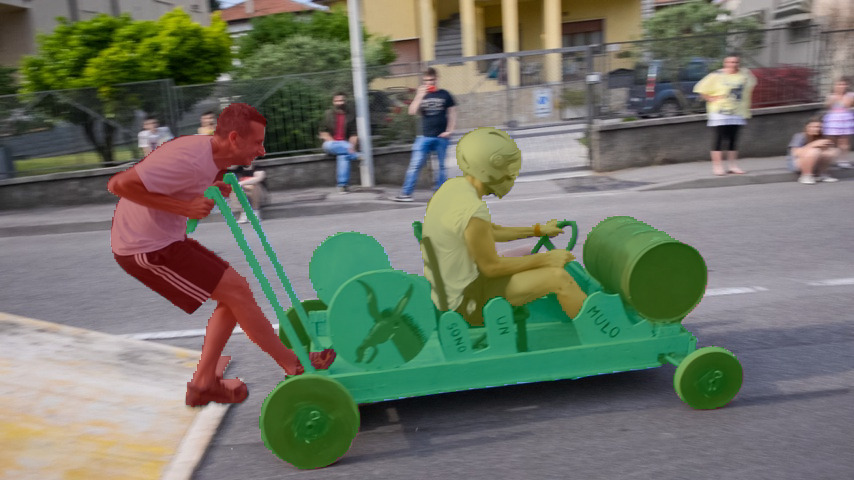}
    \end{subfigure}
    \vspace*{-0.2cm}
    \caption{\label{fig:main_case_davis} Predictions on DAVIS \textit{val} set.}
    \vspace*{-0.2cm}
\end{figure*}

Figure~\ref{fig:main_case_coco},~\ref{fig:main_case_city} and~\ref{fig:main_case_davis} show example results of \ptsd on MS-COCO, Cityscape, and DAVIS. 
One can see that our model is capable of capturing small objects in dense scene well. More visuslizations are shown in Appnedix~\ref{app:vis}.

 \section{Related Work}

\vspace*{-0.1cm}
\paragraph{Image Panoptic Segmentation.}
Panoptic segmentation, introduced in~\cite{kirillov2019panoptic}, unifies semantic segmentation and instance segmentation. Previous approaches to panoptic segmentation involve pipelines with multiple stages, such as object detection, semantic and instance segmentation, and the fusion of separate predictions~\cite{kirillov2019panoptic,li2019attention,xiong2019upsnet,liu2019end,cheng2020panoptic,carion2020end}. With multiple stages involved, learning is often not  end-to-end.
Recent work has proposed end-to-end approaches with Transformer based architectures~\cite{wang2021max,cheng2021per,zhang2021k,cheng2021masked,li2022panoptic,yu2022cmt,yu2022kmeans,li2022mask}, for which the model directly predicts segmentation masks and optimizes based on a bipartite graph matching loss. Nevertheless, they still require customized architectures (e.g., per instance mask generation, and mask fusion module). Their loss functions are also specialized for optimizing metrics used in object matching.

Our approach is a significant departure to existing methods with task-specific designs, as we simply treat the task as image-conditional discrete mask generation, without reliance on inductive biases of the task. This results in a simpler and more generic design, one which is easily extended to video segmentation with minimal modifications.

\vspace*{-0.2cm}
\paragraph{Video Segmentation.}
Among the numerous video segmentation tasks, video object segmentation (VOS) \cite{perazzi2016benchmark, wang2021survey} is perhaps the most canonical task, which entails the segmentation of key objects (of unknown categories). Video instance segmentation (VIS) \cite{yang2019video} is similar to VOS, but requires category prediction of instances. 
Video panoptic segmentation (VPS) \cite{kim2020video} is a direct extension of image panoptic segmentation to the video domain. All video segmentation tasks involve two main challenges, namely, segmentation and object tracking. And like image segmentation, most existing methods are specialist models comprising multiple stages with pipe-lined frameworks, e.g., track-detect \cite{yang2019video, voigtlaender2019mots, lin2020video, cao2020sipmask, Luiten_2020_WACV}, clip-match \cite{athar2020stem, bertasius2020classifying}, propose-reduce \cite{Lin_2021_ICCV}. End-to-end approaches have also been proposed recently \cite{wang2021end,kim2022tubeformer}, but with a specialized loss function.

In this work we directly take the \ptsd model, pretrained on COCO for panoptic segmentation, and fine-tune it for unsupervised video object segmentation (UVOS), where it performs VOS without the need for manual initialization. Model architectures, training losses, input augmentations and sampling methods all remained largely unchanged when applied to UVOS data. Because of this, we believe it is just as straightforward to apply \ptsd to the other video segmentation tasks as well.

\vspace*{-0.2cm}
\paragraph{Others.} Our work is also related to recent generalist vision models~\cite{chen2021pix2seq,chen2022unified,wang2022unifying,UViM,unifiedIO} where both architecture and loss functions are task-agnostic. Existing generalist models are based on autoregressive models, while our work is based on Bit Diffusion~\cite{chen2022analog,ho2020denoising,sohl2015deep,song2020denoising}. 
Diffusion models have been applied to semantic segmentation, directly~\cite{amit2021segdiff,wolleb2021diffusion,kim2022diffusion} or indirectly~\cite{baranchuk2021label,Asiedu2022-CVPRW}.
However none of these methods model segmentation masks as discrete/categorical tokens, nor are their models capable of video segmentation.
 \section{Conclusion and Future Work}

This paper proposes a simple framework for panoptic segmentation of images and videos, based on conditional generative models of discrete panoptic masks. 
Our approach is able to model large numbers of discrete tokens ($10^6$ in our experiments), which is difficult with other existing generative segmentation models.
We believe both the architecture, modeling choices, and training procedure (including augmentations) we use here can be further improved to boost the performance.
Furthermore, the required inference steps can also be further reduced with techniques like progressive distillation.
Finally, we want to note that, as a significant departure to status quo, we acknowledge that our current empirical result is still behind compared to well-tuned pipelines in existing systems  (though our results are still competitive and at a practically usable level). However, with the simplicity of the proposed approach, we hope it would spark future development that drives new state-of-the-art systems.

\section*{Acknowledgements}
We specially thank Liang-Chieh Chen for the helpful feedback on our initial draft.

{\small
\bibliographystyle{ieee_fullname}
\bibliography{content/ref}
}

\newpage
\clearpage
\appendix

\section{Algorithm details}
\label{app:alg}

For completeness, we include Algorithm~\ref{alg:encode_decode} and~\ref{alg:ddim_ddpm} from Bit Diffusion~\cite{chen2022analog} to provide more detailed implementations of functions in Algorithm~\ref{alg:train} and~\ref{alg:sample}.

\begin{algorithm}[H]
\small
\caption{\small Binary encoding and decoding algorithms (in Tensorflow).
}
\label{alg:encode_decode}
\definecolor{codeblue}{rgb}{0.25,0.5,0.5}
\definecolor{codekw}{rgb}{0.85, 0.18, 0.50}
\lstset{
  backgroundcolor=\color{white},
  basicstyle=\fontsize{7.5pt}{7.5pt}\ttfamily\selectfont,
  columns=fullflexible,
  breaklines=true,
  captionpos=b,
  commentstyle=\fontsize{7.5pt}{7.5pt}\color{codeblue},
  keywordstyle=\fontsize{7.5pt}{7.5pt}\color{codekw},
  escapechar={|}, 
}
\begin{lstlisting}[language=python]
import tensorflow as tf

def int2bit(x, n=8):
  # Convert integers into corresponding binary bits.
  x = tf.bitwise.right_shift(
        tf.expand_dims(x, -1), tf.range(n))
  x = tf.math.mod(x, 2)
  return x
    
def bit2int(x):
  # Convert binary bits into corresponding integers.
  x = tf.cast(x, tf.int32)
  n = x.shape[-1]
  x = tf.math.reduce_sum(x * (2 ** tf.range(n)), -1)
  return x
\end{lstlisting}
\end{algorithm}

\begin{algorithm}[H]
\small
\caption{\small $x_t$ estimation with DDIM updating rule.
}
\label{alg:ddim_ddpm}
\definecolor{codeblue}{rgb}{0.25,0.5,0.5}
\definecolor{codekw}{rgb}{0.85, 0.18, 0.50}
\lstset{
  backgroundcolor=\color{white},
  basicstyle=\fontsize{7.5pt}{7.5pt}\ttfamily\selectfont,
  columns=fullflexible,
  breaklines=true,
  captionpos=b,
  commentstyle=\fontsize{7.5pt}{7.5pt}\color{codeblue},
  keywordstyle=\fontsize{7.5pt}{7.5pt}\color{codekw},
  escapechar={|}, 
}
\begin{lstlisting}[language=python]
def gamma(t, ns=0.0002, ds=0.00025):
  # A scheduling function based on cosine function.
  return numpy.cos(
        ((t + ns) / (1 + ds)) * numpy.pi / 2)**2

def ddim_step(x_t, x_pred, t_now, t_next):
  # Estimate x at t_next with DDIM updating rule.
  |$\gamma_{\text{now}}$| = gamma(t_now)
  |$\gamma_{\text{next}}$| = gamma(t_next)
  x_pred = clip(x_pred, -scale, scale)
  eps =|$\frac{1}{\sqrt{1-\gamma_\text{now}}}$| * (x_t - |$\sqrt{\gamma_\text{now}}$| * x_pred)
  x_next = |$\sqrt{\gamma_\text{next}}$| * x_pred + |$\sqrt{1-\gamma_\text{next}}$| * eps
  return x_next
\end{lstlisting}
\end{algorithm}

\section{More details on training and inference hyper-parameters}
\label{app:hparams}

\textbf{MS-COCO.} For unconditional pretraining of the mask decoder, we train the model on mask resolution 128$\times$128 for 800 epochs on MS-COCO with a batch size of 512 and scale jittering of strength $[1.0, 3.0]$. For both unconditional pretraining of the mask decoder, and image-conditional training of mask generation (encoder and decoder), we use input scaling of $0.1$, loss weighting $p=0.2$, learning rate of $1e^{-4}$, with EMA decay of 0.999.

\textbf{Cityscapes.} For fine-tuning on Cityscapes, we train for 800 epochs using a batch size of 16 and learning rate of $3e^{-5}$ linearly decayed to $3e^{-6}$ with no warmup and no scale jittering augmentation. Image size of 1024$\times$2048 with mask size of 512$\times$1024 are used. During eval, we use \verb|td|$=1.5$ and resize the predicted mask to the full mask size of $1024\times2048$ using nearest neighbour resizing and filter out annotations with less than 80 pixels. 

\textbf{DAVIS.} For fine-tuning on DAVIS, we train for 20k steps with batch size of 32, loss weighting of $p=0.2$, a constant learning rate of $1e^{-5}$, EMA decay of 0.99, and scale jittering of strength $[0.7, 2]$. Image size of 512$\times$1024 and mask size of 256$\times$512 are used. For evaluation, as the dataset is quite small, we run inference 50 times for our model, and report the mean (the standard deviation for $\mathcal{J\&F}$ is around 1.5).

\vspace{-0.2cm}
\section{Results on Cityscape}
\label{app:results_cityscape}
Table~\ref{tab:cityscapes_val} compares our results on Cityscapes \textit{val} set with prior work. Our main results are with an image size of $1024\times2048$ and mask size of $512\times1024$. In Table~\ref{tab:cityscapes_image_size_ablation} we show an ablation with varying image and mask sizes and we find that both a larger image size and a larger mask size are important.

\begin{table}[!t]
\centering
\small
\begin{tabular}{r|c|c|c}
\toprule
method & backbone & params & PQ \\
\midrule
\multicolumn{4}{l}{\textit{Specialist approaches:\hfill}} \\
Mask2Former~\cite{cheng2021masked} & ResNet-50~\cite{he2016deep} & - & 62.1 \\
kMaX-DeepLab~\cite{yu2022kmeans} & ResNet-50~\cite{he2016deep} & 56M & 64.3 \\
\hline
Panoptic-DeepLab~\cite{cheng2019panoptic} & Xception-71~\cite{chollet2016xception} & 47M & 63.0 \\
Axial-DeepLab~\cite{wang2020axial} & Axial-ResNet-L~\cite{wang2020axial} & 45M & 63.9 \\
Axial-DeepLab~\cite{wang2020axial} & Axial-ResNet-XL~\cite{wang2020axial} & 173M & 64.4 \\
CMT-DeepLab~\cite{yu2022cmt} & MaX-S~\cite{wang2021max} & - & 64.6 \\
Panoptic-DeepLab~\cite{cheng2019panoptic} & SWideRNet-(1,1,4.5)~\cite{swidernet_2020} & 536M & 66.4\\
Mask2Former~\cite{cheng2021masked} & Swin-B (W12)~\cite{liu2021swin} & - & 66.1 \\
Mask2Former~\cite{cheng2021masked} & Swin-L (W12)~\cite{liu2021swin} & - & 66.6 \\
kMaX-DeepLab~\cite{yu2022kmeans} & MaX-S~\cite{wang2021max} & 74M & 66.4 \\
kMaX-DeepLab~\cite{yu2022kmeans} & ConvNeXt-B~\cite{liu2022convnet} & 121M & 68.0 \\
kMaX-DeepLab~\cite{yu2022kmeans} & ConvNeXt-L~\cite{liu2022convnet} & 232M & 68.4 \\
\hline
\multicolumn{4}{l}{\textit{Generalist approaches:\hfill}} \\
\ptsd (steps=10) & ResNet-50~\cite{he2016deep} & 94.8M & 62.2 \\
\ptsd (steps=20) & ResNet-50~\cite{he2016deep} & 94.8M & 63.2 \\
\ptsd (steps=40) & ResNet-50~\cite{he2016deep} & 94.8M & 63.4 \\
\ptsd (steps=80) & ResNet-50~\cite{he2016deep} & 94.8M & 64.0 \\
\bottomrule
\end{tabular}
\vspace*{-0.2cm}
\caption{Cityscapes \textit{val} set results.}
\vspace*{-0.2cm}
\label{tab:cityscapes_val}
\end{table}

\begin{table}[!t]
\centering
\small
\begin{tabular}{r|c|c}
\toprule
&\multicolumn{2}{c}{Mask size} \\
Image size & $256\times512$ &  $512\times1024$\\
\midrule
 $512\times1024$ & 52.1 & 53.8 \\
$1024\times2048$ & 55.7 & 59.9 \\
\bottomrule
\end{tabular}
\vspace*{-0.2cm}
\caption{PQ for various image sizes  and panoptic mask sizes for Cityscapes \textit{val} set. Model is trained for 100 epochs for ablation.}
\vspace*{-0.5cm}
\label{tab:cityscapes_image_size_ablation}
\end{table}

\vspace{-0.2cm}
\section{Results on KITTI-STEP}

In addition to video object segmentation on DAVIS, we also applied the same method to video panoptic segmentation on more recent KITTI-STEP dataset \cite{weber2021step}. Training configurations remain the same as on DAVIS, but with image size 384x1248. For inference we use $td=1.5$ and 10 sampling steps. Our preliminary results are shown in Table~\ref{tab:kittistep}. Pix2seq-D achieved decent results (though behind the state-of-the-art), especially considering that minimal tuning or changes are done to apply Pix2seq-D to this task.

\begin{table}[h]
\centering
\small
\begin{tabular}{llll}
\toprule
Method & STQ & SQ & AQ \\
\midrule
Motion-DeepLab \cite{weber2021step} & 58.0 & 67.0 & 51.0 \\ VPSNet \cite{kim2020video} & 56.0 & 61.0 & 52.0 \\ TubeFormer-DeepLab-B1 \cite{kim2022tubeformer} & 70.0 & 76.8 & 63.8 \\ Pix2Seq-D & 61.6 & 61.9 & 61.3 \\
\bottomrule
\end{tabular}
\vspace*{-0.2cm}
\caption{Comparison of results on KITTI-STEP. All methods listed have a Resnet-50 backbone.}
\vspace*{-0.5cm}
\label{tab:kittistep}
\end{table}

Unlike most existing methods that do inference on video segments and stitch the segments in postprocessing, we do inference on the entire video in a streaming fashion. For the ease of experimentation, we only conduct inference on videos up to 400 frames with image size 384x1248. One video in the KITTI-STEP validation set exceeds 400 frames, and is therefore split into two videos during inference. We believe this only has a negligible impact on the overall metrics.

\vspace{-0.2cm}
\section{Extra Visualization}
\label{app:vis}

\begin{figure*}[t]
    \centering
    \begin{subfigure}{0.9\textwidth}
      \centering
      \includegraphics[width=\linewidth]{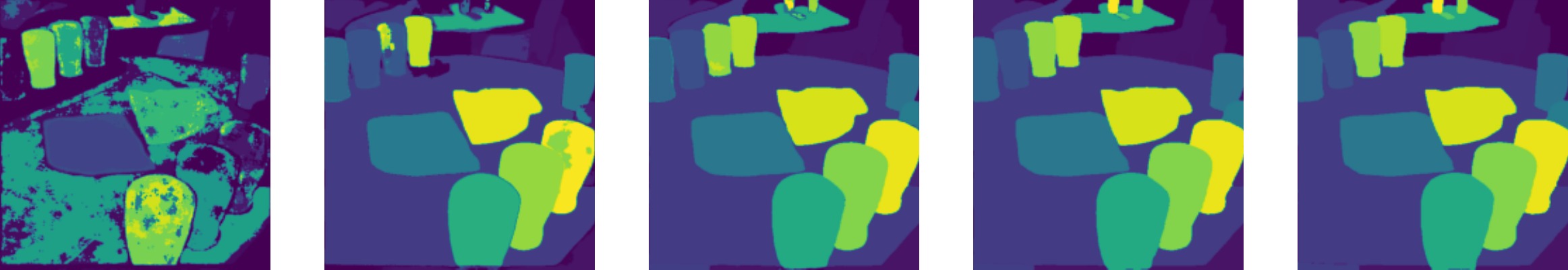}
    \end{subfigure}
    \begin{subfigure}{0.9\textwidth}
      \centering
      \includegraphics[width=\linewidth]{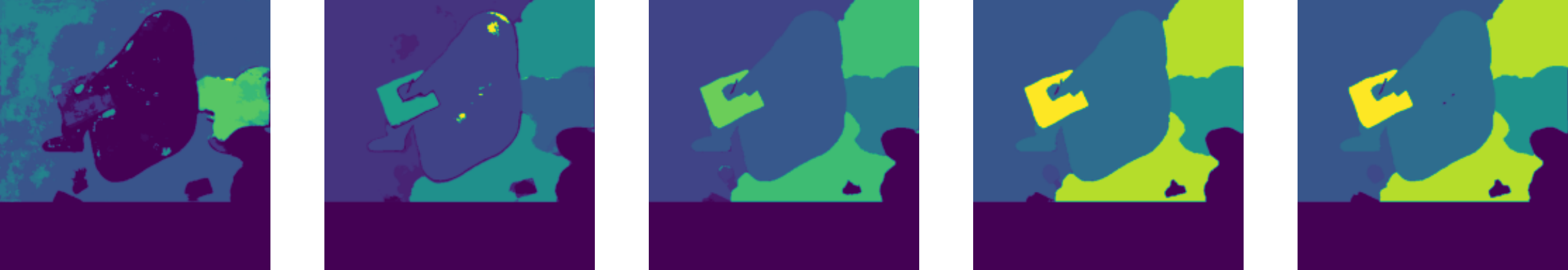}
    \end{subfigure}
    \caption{\label{fig:inf_traj} Inference trajectory. Predicted $m_0$ at different time steps (1, 2, 4, 8, 16) out of total 20 steps.}
\end{figure*}

Figure~\ref{fig:inf_traj} shows the inference trajectory of or model for two MS-COCO validation examples, we see that the model iteratively refines the panoptic mask outputs so that they become globally and locally consistent.

\begin{figure*}[t]
    \centering
    \begin{subfigure}{0.32\textwidth}
      \centering
      Image
    \end{subfigure}
    \begin{subfigure}{0.32\textwidth}
      \centering
      Prediction
    \end{subfigure}
    \begin{subfigure}{0.32\textwidth}
      \centering
      Groundtruth
    \end{subfigure}
    
    \begin{subfigure}{0.3\textwidth}
      \centering
      \includegraphics[width=\linewidth]{figures/case_000000001000_img.jpeg}
    \end{subfigure}
    \begin{subfigure}{0.3\textwidth}
      \centering
      \includegraphics[width=\linewidth]{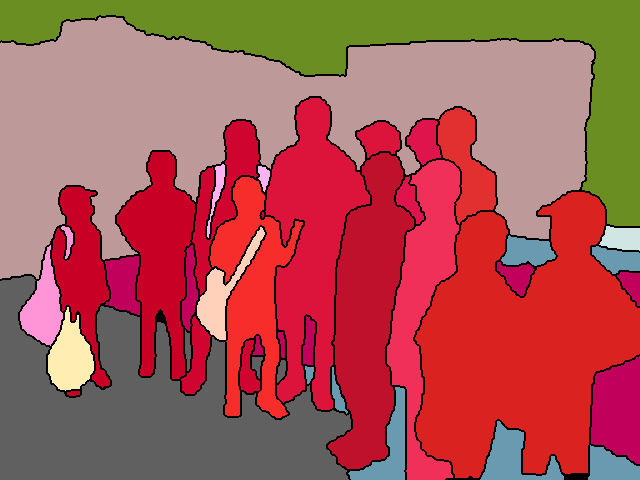}
    \end{subfigure}
    \begin{subfigure}{0.3\textwidth}
      \centering
      \includegraphics[width=\linewidth]{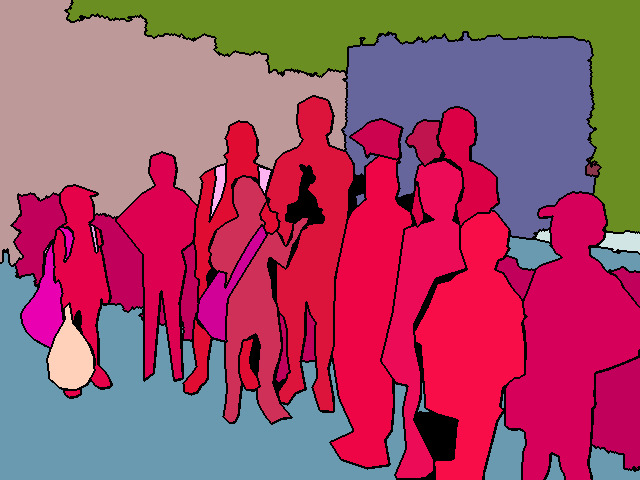}
    \end{subfigure}
    \begin{subfigure}{0.3\textwidth}
      \centering
      \includegraphics[width=\linewidth]{figures/case_000000001584_img.jpeg}
    \end{subfigure}
    \begin{subfigure}{0.3\textwidth}
      \centering
      \includegraphics[width=\linewidth]{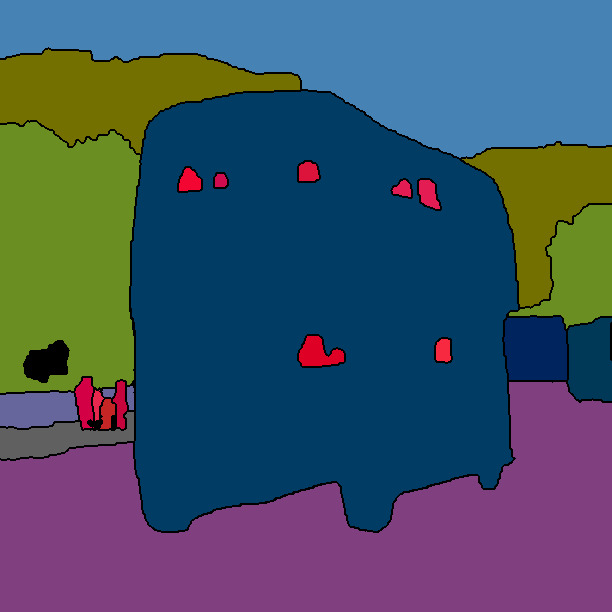}
    \end{subfigure}
    \begin{subfigure}{0.3\textwidth}
      \centering
      \includegraphics[width=\linewidth]{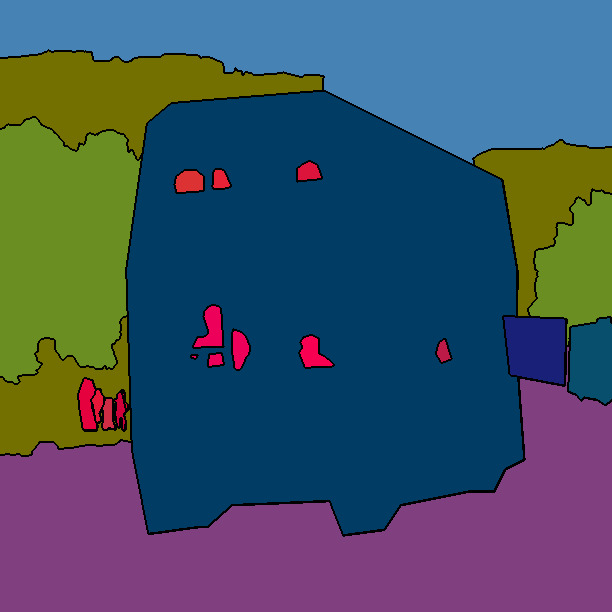}
    \end{subfigure}
    \begin{subfigure}{0.3\textwidth}
      \centering
      \includegraphics[width=\linewidth]{figures/case_000000001268_img.jpeg}
    \end{subfigure}
    \begin{subfigure}{0.3\textwidth}
      \centering
      \includegraphics[width=\linewidth]{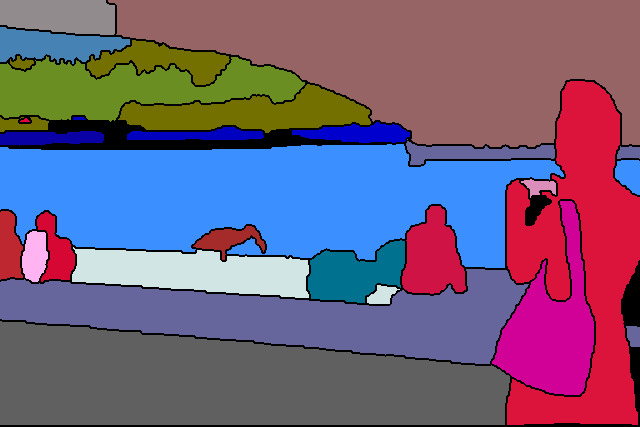}
    \end{subfigure}
    \begin{subfigure}{0.3\textwidth}
      \centering
      \includegraphics[width=\linewidth]{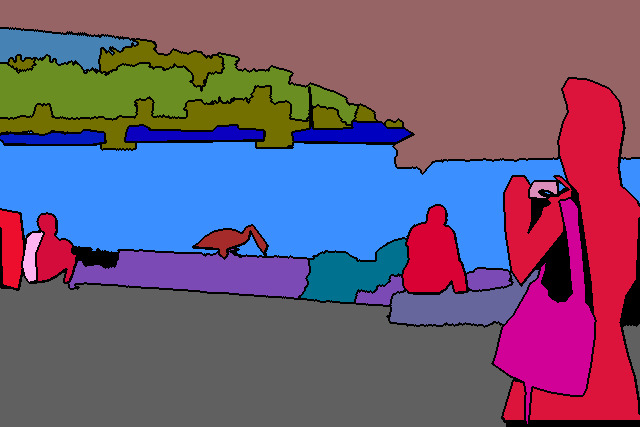}
    \end{subfigure}
\caption{\label{fig:more_case_coco1} Predictions on MS-COCO \textit{val} set.}
\end{figure*}

\begin{figure*}[t]
    \centering
    \begin{subfigure}{0.32\textwidth}
      \centering
      Image
    \end{subfigure}
    \begin{subfigure}{0.32\textwidth}
      \centering
      Prediction
    \end{subfigure}
    \begin{subfigure}{0.32\textwidth}
      \centering
      Groundtruth
    \end{subfigure}
    
    \begin{subfigure}{0.3\textwidth}
      \centering
      \includegraphics[width=\linewidth]{figures/case_000000272148_img.jpeg}
    \end{subfigure}
    \begin{subfigure}{0.3\textwidth}
      \centering
      \includegraphics[width=\linewidth]{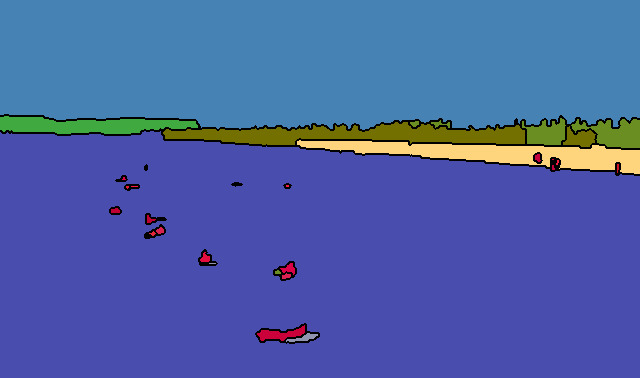}
    \end{subfigure}
    \begin{subfigure}{0.3\textwidth}
      \centering
      \includegraphics[width=\linewidth]{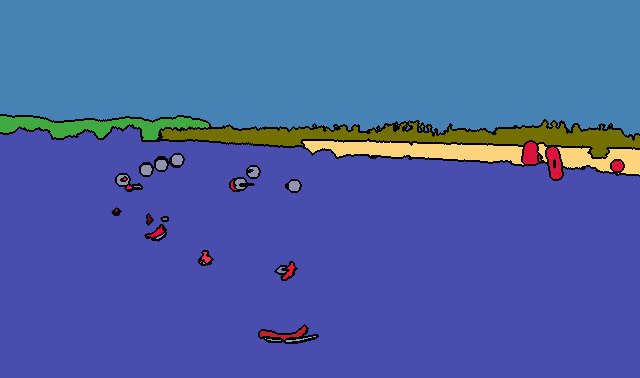}
    \end{subfigure}
    \begin{subfigure}{0.3\textwidth}
      \centering
      \includegraphics[width=\linewidth]{figures/case_000000309391_img.jpeg}
    \end{subfigure}
    \begin{subfigure}{0.3\textwidth}
      \centering
      \includegraphics[width=\linewidth]{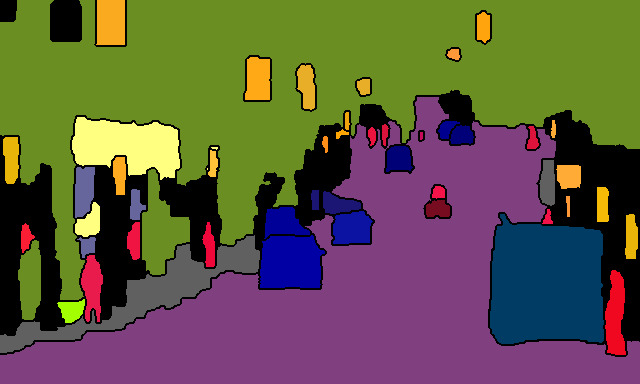}
    \end{subfigure}
    \begin{subfigure}{0.3\textwidth}
      \centering
      \includegraphics[width=\linewidth]{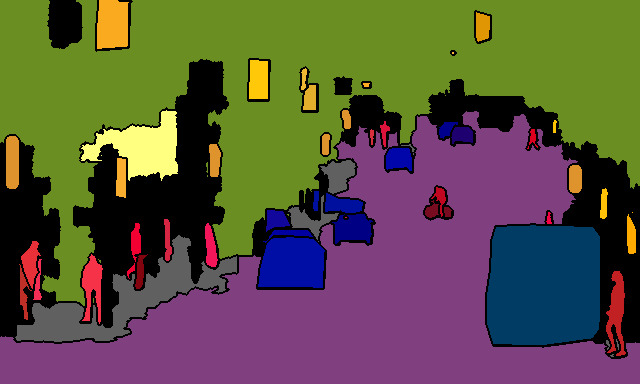}
    \end{subfigure}
    \begin{subfigure}{0.3\textwidth}
      \centering
      \includegraphics[width=\linewidth]{figures/case_000000432898_img.jpeg}
    \end{subfigure}
    \begin{subfigure}{0.3\textwidth}
      \centering
      \includegraphics[width=\linewidth]{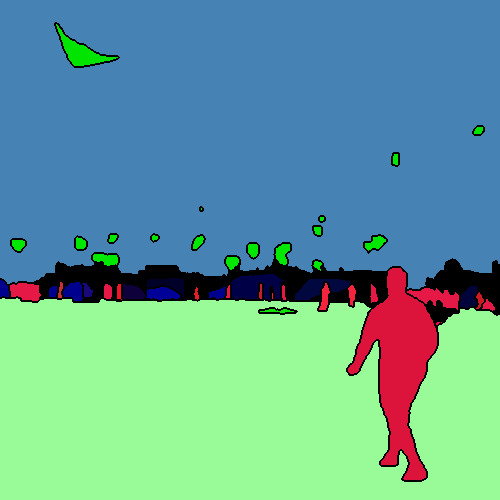}
    \end{subfigure}
    \begin{subfigure}{0.3\textwidth}
      \centering
      \includegraphics[width=\linewidth]{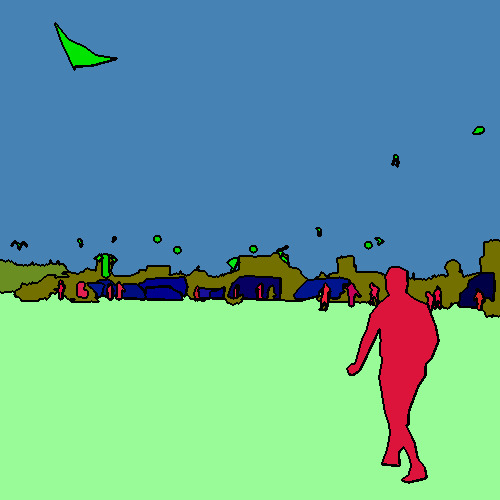}
    \end{subfigure}
    \begin{subfigure}{0.3\textwidth}
      \centering
      \includegraphics[width=\linewidth]{figures/case_000000439180_img.jpeg}
    \end{subfigure}
    \begin{subfigure}{0.3\textwidth}
      \centering
      \includegraphics[width=\linewidth]{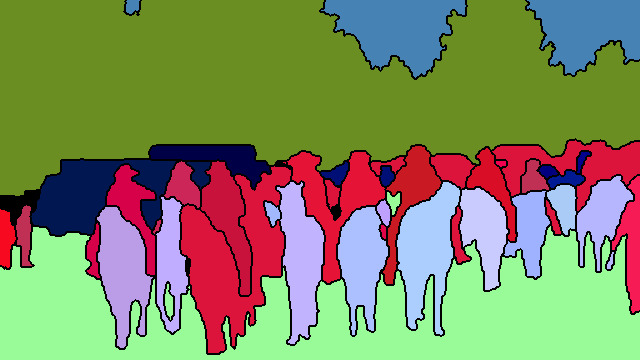}
    \end{subfigure}
    \begin{subfigure}{0.3\textwidth}
      \centering
      \includegraphics[width=\linewidth]{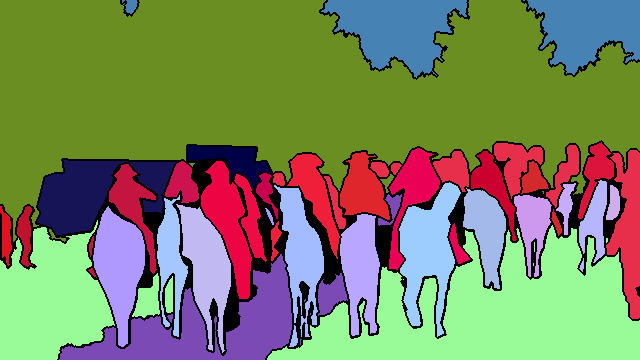}
    \end{subfigure}
    \begin{subfigure}{0.3\textwidth}
      \centering
      \includegraphics[width=\linewidth]{figures/case_000000441553_img.jpeg}
    \end{subfigure}
    \begin{subfigure}{0.3\textwidth}
      \centering
      \includegraphics[width=\linewidth]{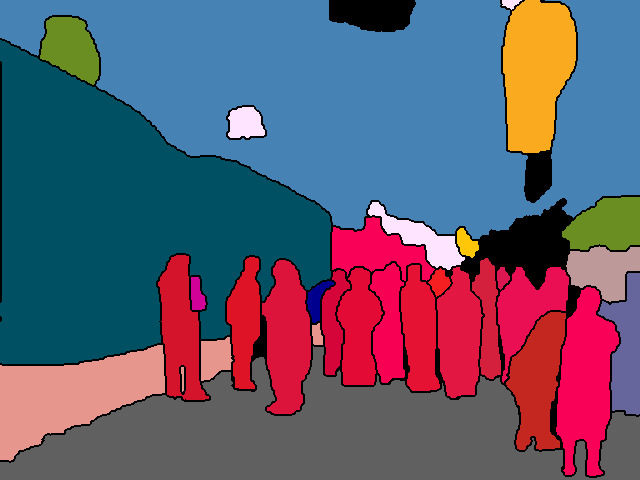}
    \end{subfigure}
    \begin{subfigure}{0.3\textwidth}
      \centering
      \includegraphics[width=\linewidth]{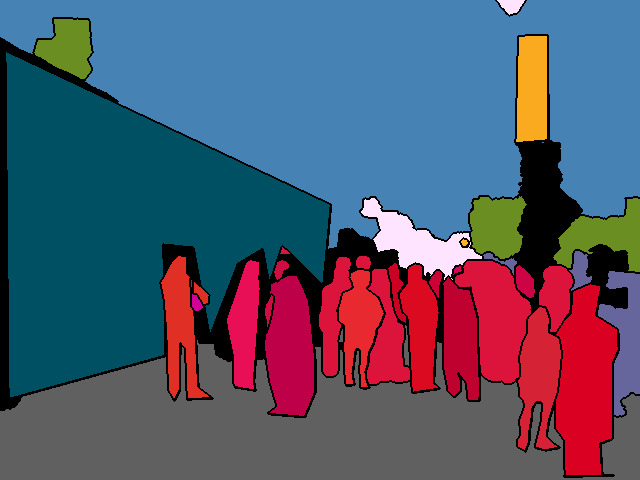}
    \end{subfigure}
    \caption{\label{fig:more_case_coco2} Predictions on MS-COCO \textit{val} set.}
\end{figure*}

\begin{figure*}[t]
    \centering
    \begin{subfigure}{0.32\textwidth}
      \centering
      Image
    \end{subfigure}
    \begin{subfigure}{0.32\textwidth}
      \centering
      Prediction
    \end{subfigure}
    \begin{subfigure}{0.32\textwidth}
      \centering
      Groundtruth
    \end{subfigure}

    \begin{subfigure}{0.32\textwidth}
      \centering
      \includegraphics[width=\linewidth]{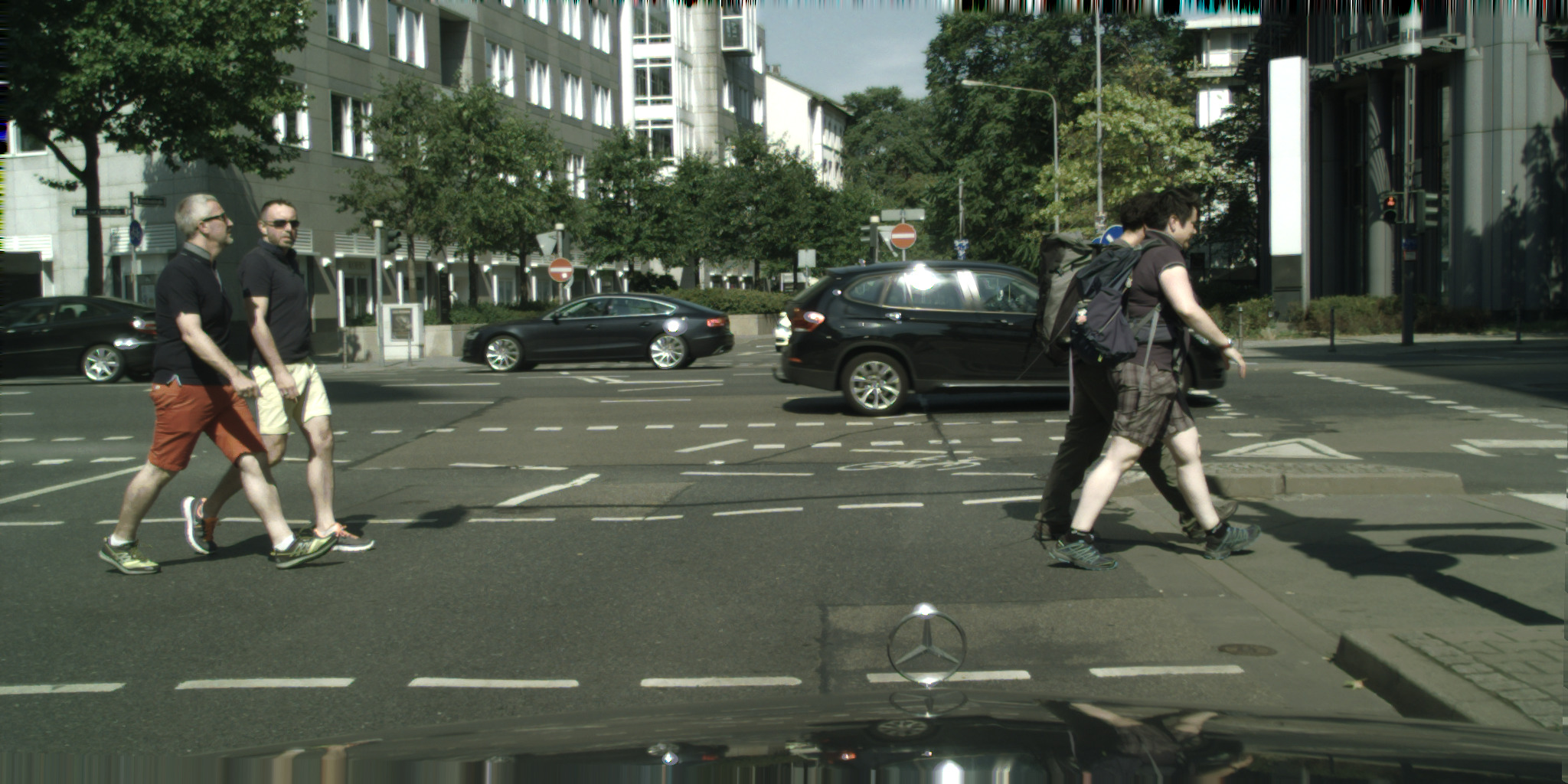}
    \end{subfigure}
    \begin{subfigure}{0.32\textwidth}
      \centering
      \includegraphics[width=\linewidth]{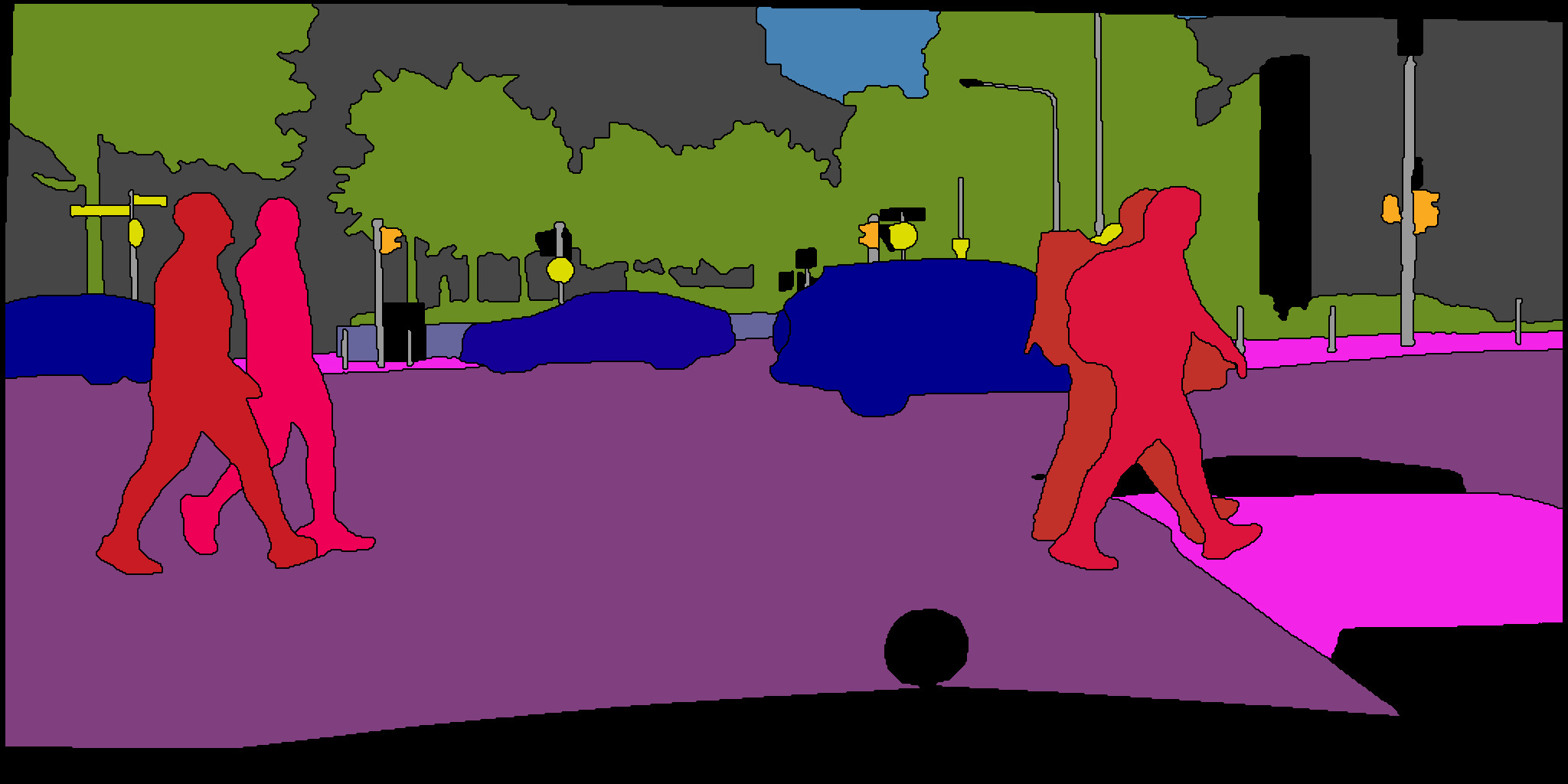}
    \end{subfigure}
    \begin{subfigure}{0.32\textwidth}
      \centering
      \includegraphics[width=\linewidth]{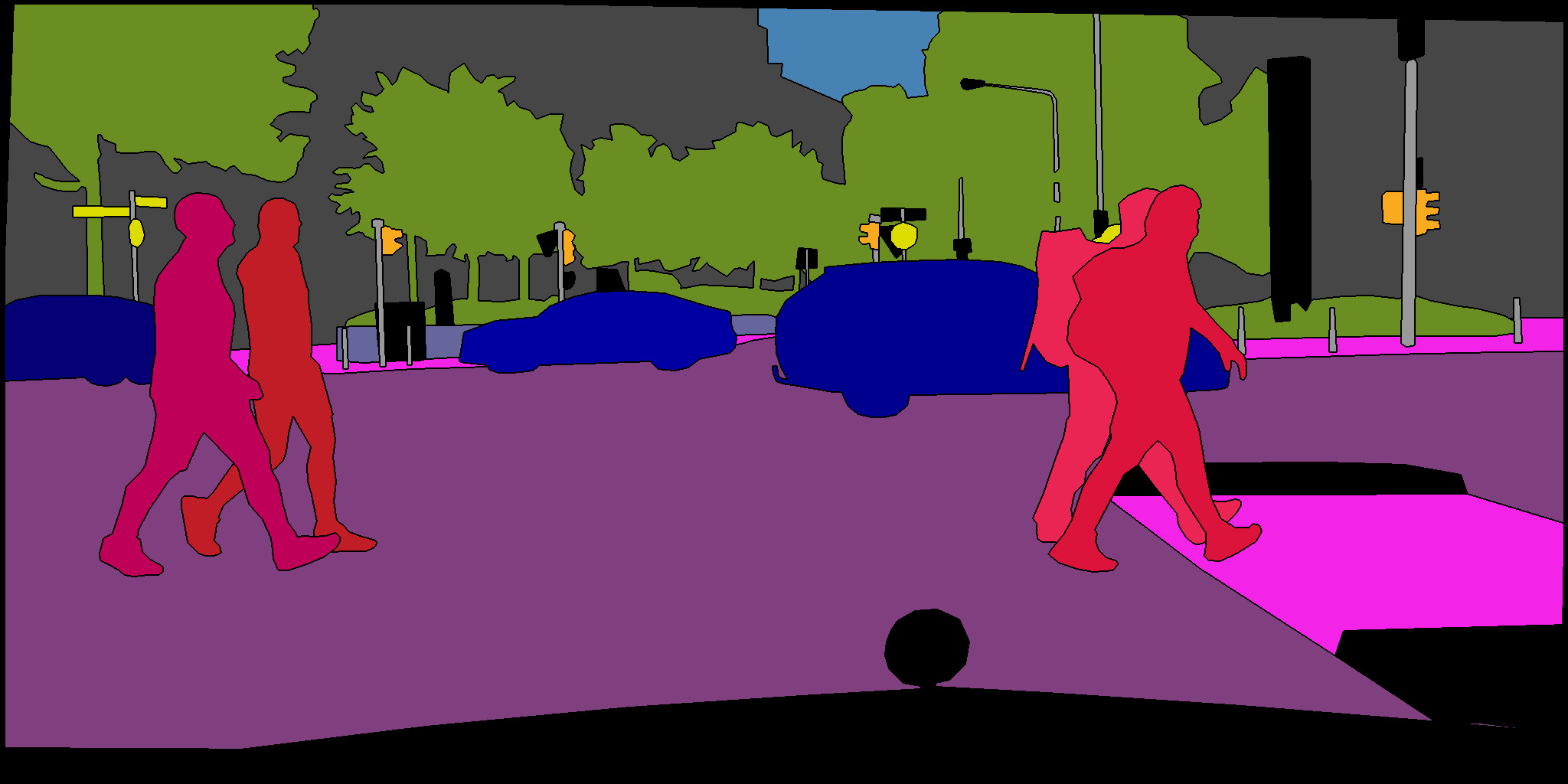}
    \end{subfigure}
    
    \begin{subfigure}{0.32\textwidth}
      \centering
      \includegraphics[width=\linewidth]{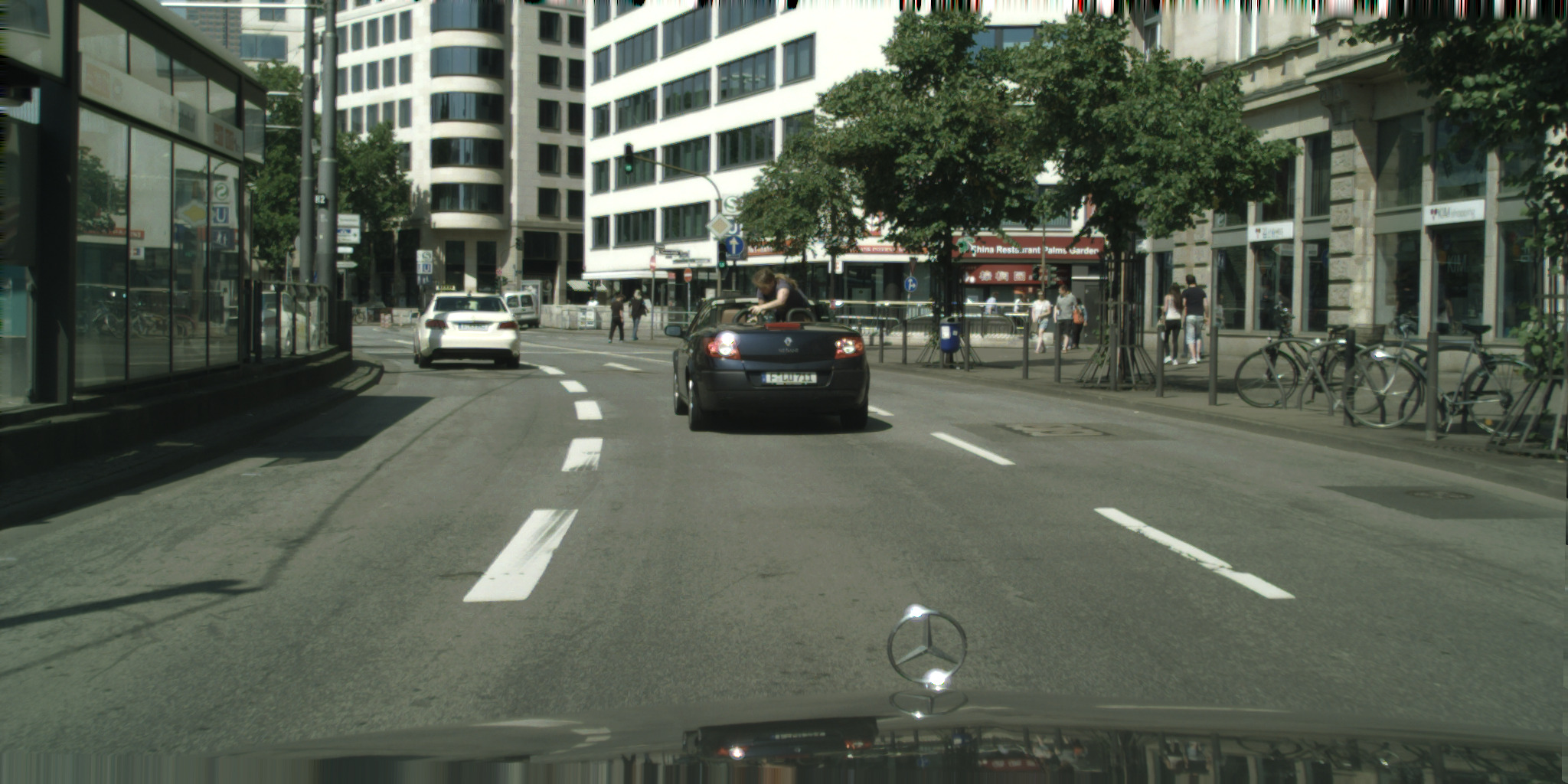}
    \end{subfigure}
    \begin{subfigure}{0.32\textwidth}
      \centering
      \includegraphics[width=\linewidth]{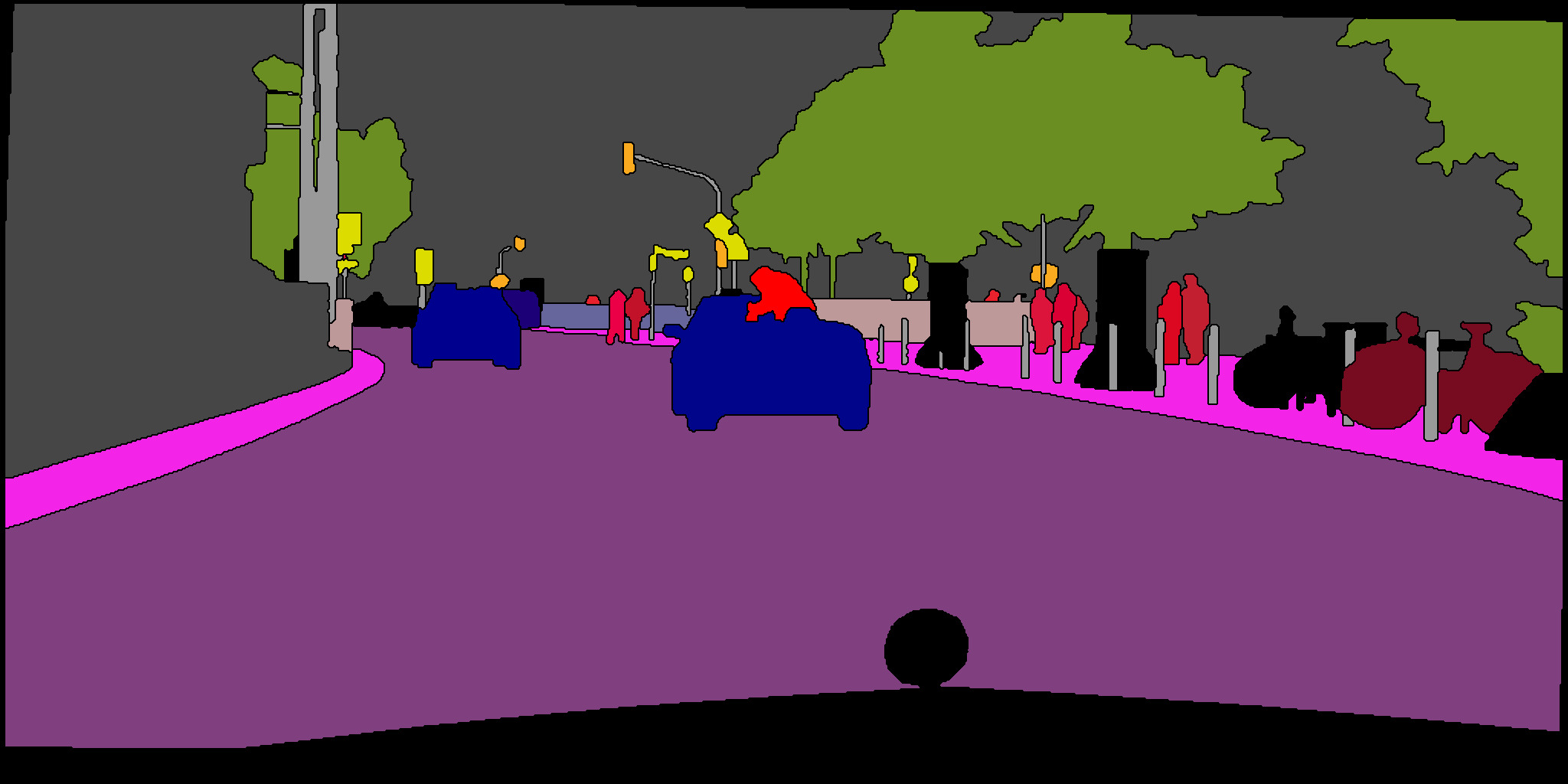}
    \end{subfigure}
    \begin{subfigure}{0.32\textwidth}
      \centering
      \includegraphics[width=\linewidth]{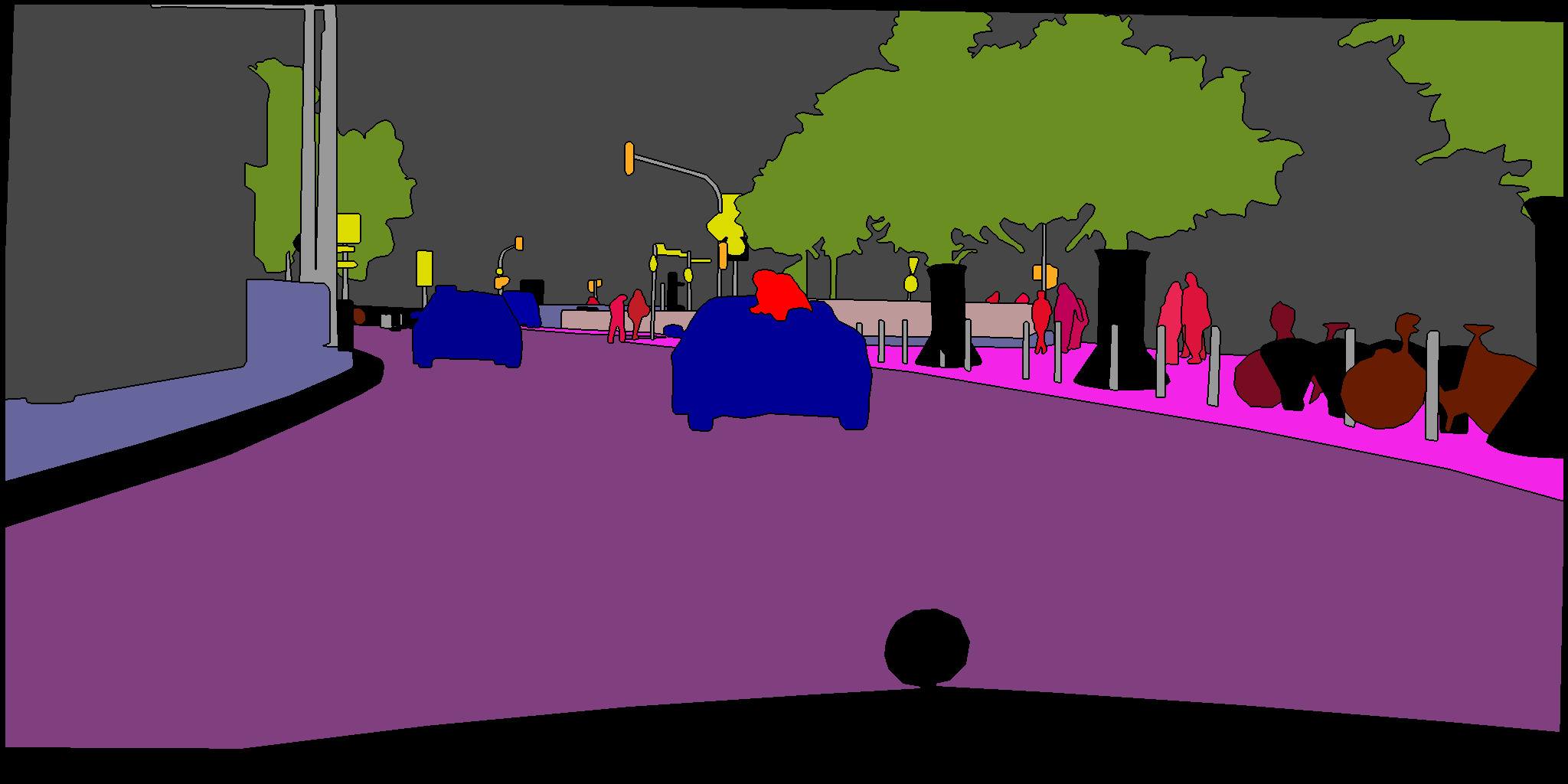}
    \end{subfigure}

    \begin{subfigure}{0.32\textwidth}
      \centering
      \includegraphics[width=\linewidth]{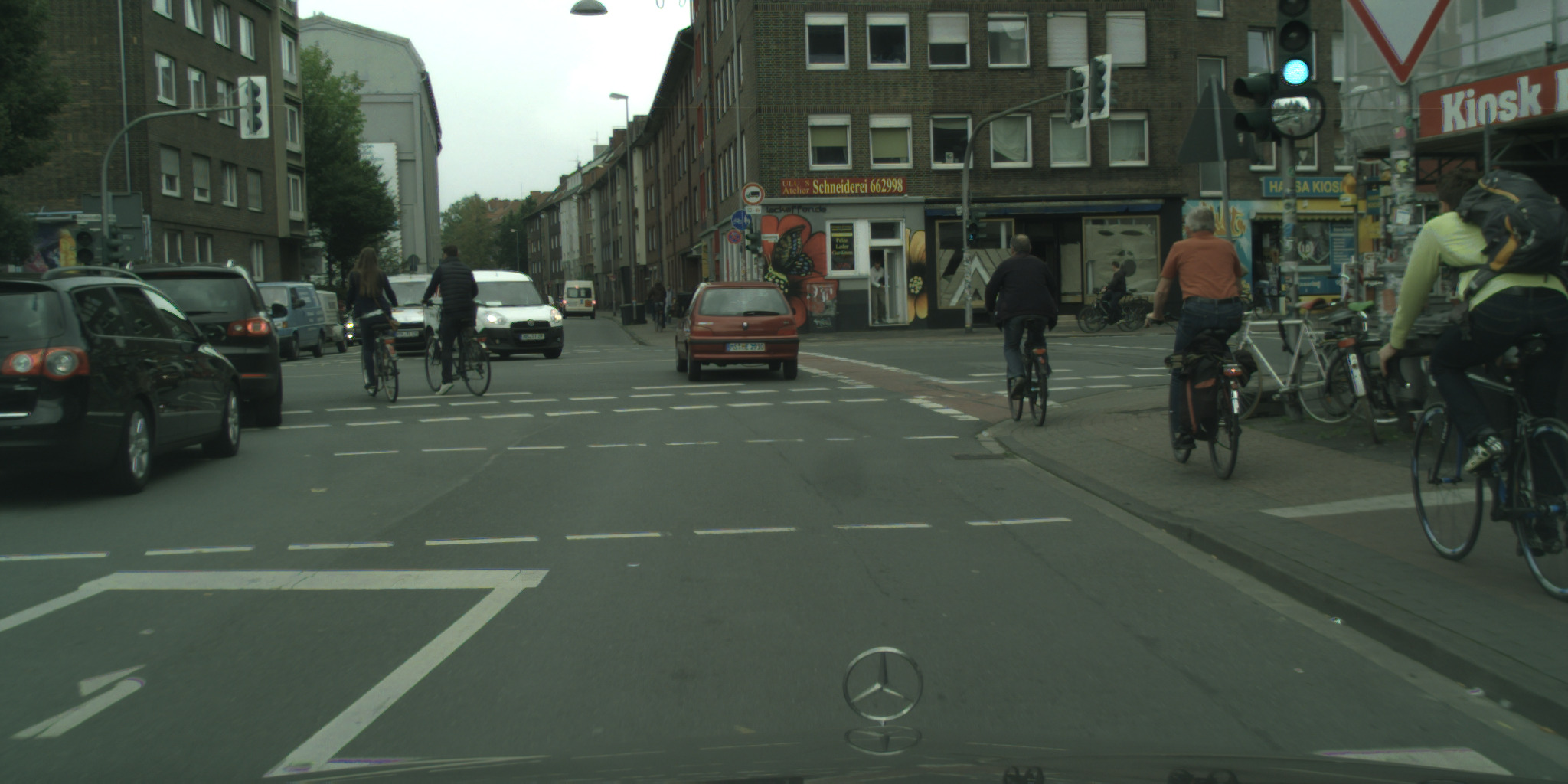}
    \end{subfigure}
    \begin{subfigure}{0.32\textwidth}
      \centering
      \includegraphics[width=\linewidth]{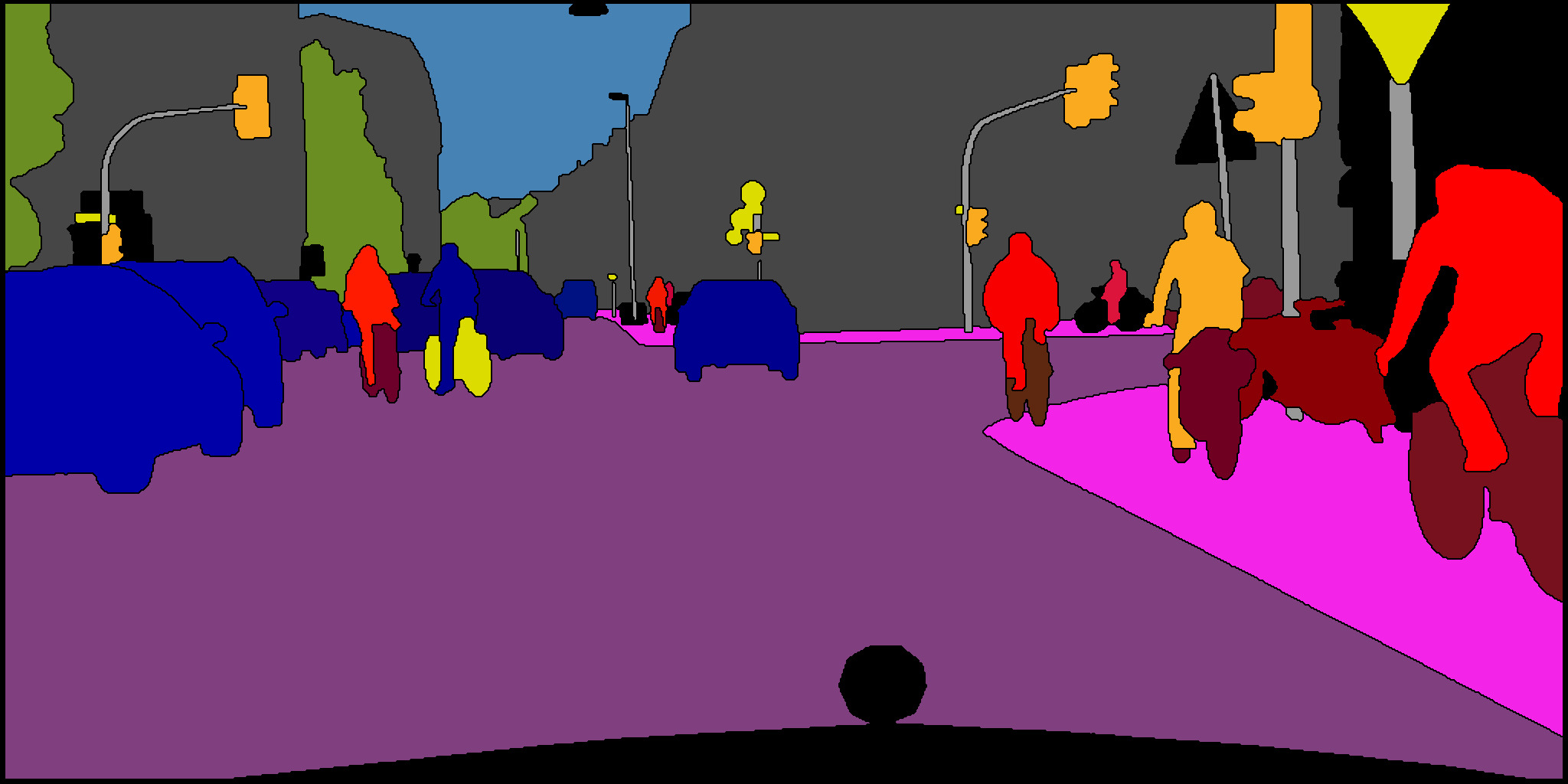}
    \end{subfigure}
    \begin{subfigure}{0.32\textwidth}
      \centering
      \includegraphics[width=\linewidth]{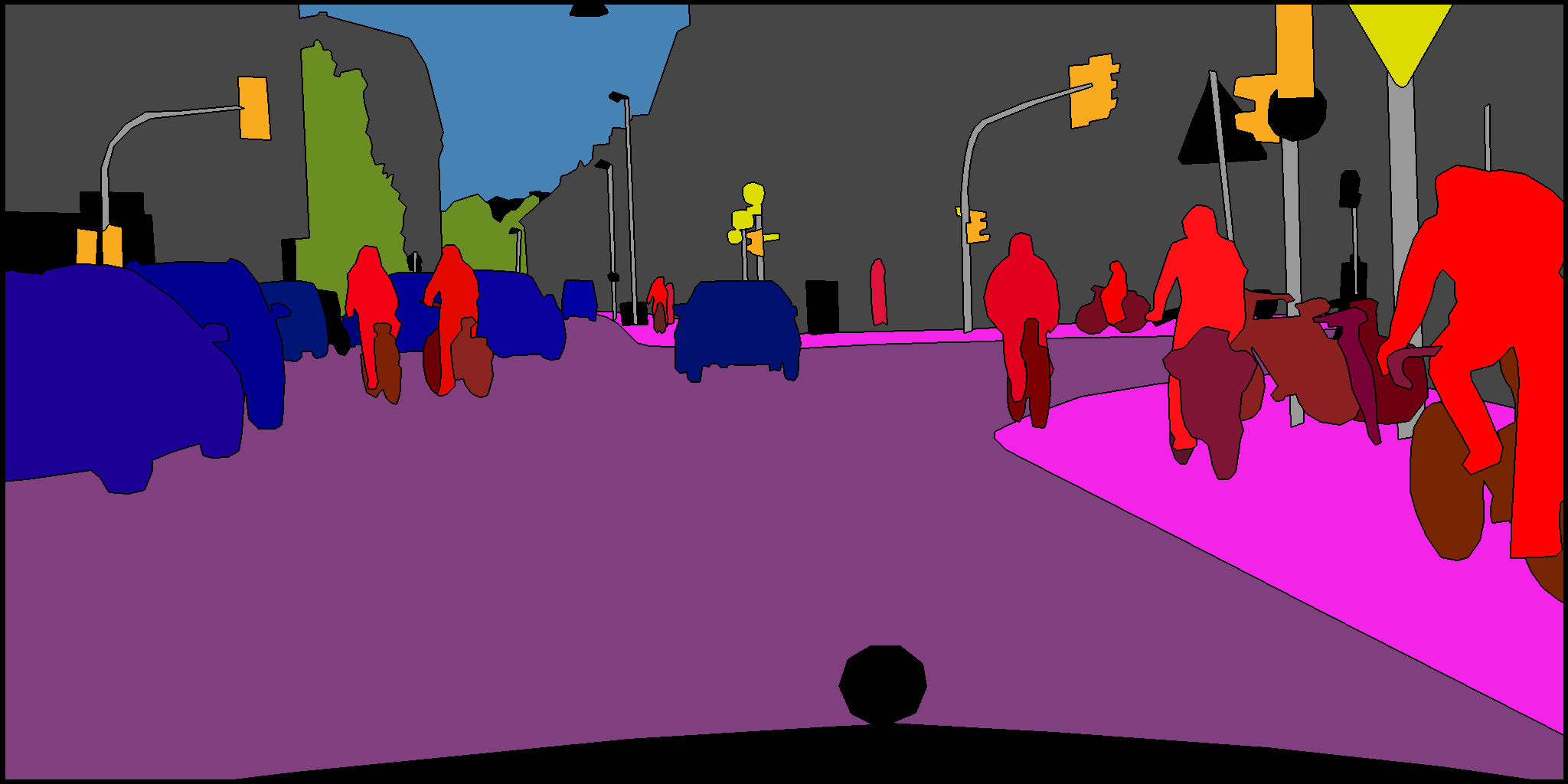}
    \end{subfigure}

    \caption{\label{fig:more_case_cityscapes} Predictions on Cityscapes \textit{val} set.}
\end{figure*}

\begin{figure*}[t]
    \centering
\begin{subfigure}{0.24\textwidth}
      \centering
      \includegraphics[width=\linewidth]{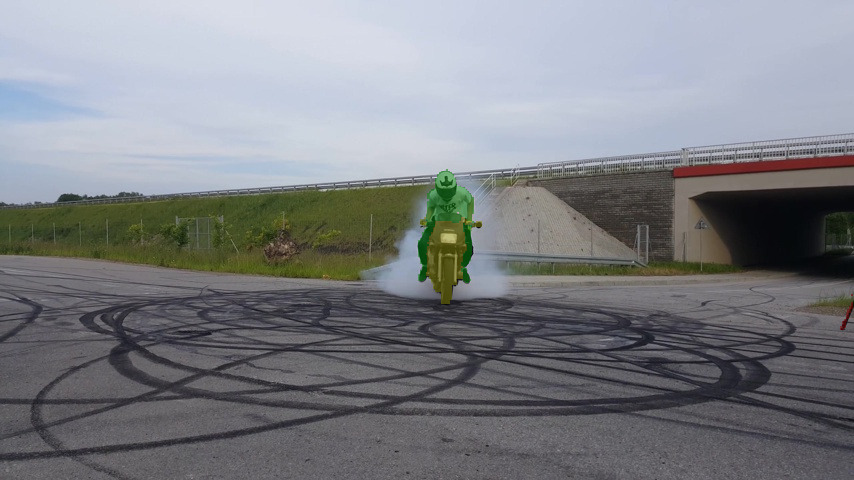}
    \end{subfigure}
    \begin{subfigure}{0.24\textwidth}
      \centering
      \includegraphics[width=\linewidth]{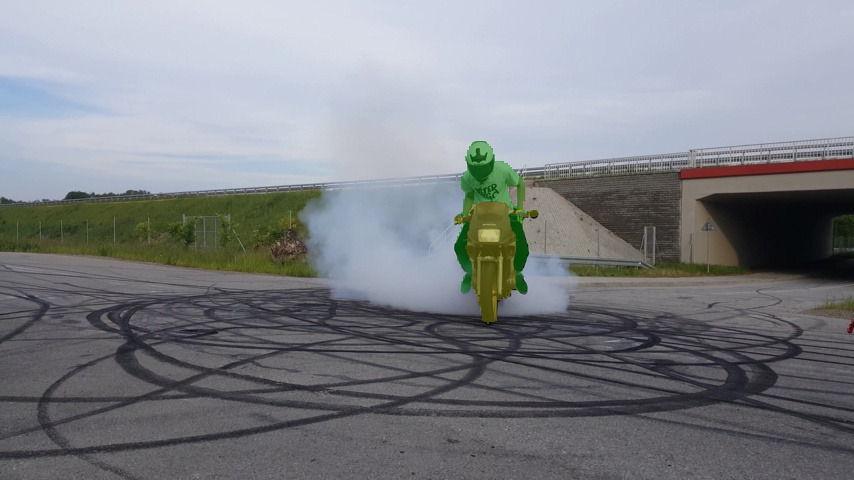}
    \end{subfigure}
    \begin{subfigure}{0.24\textwidth}
      \centering
      \includegraphics[width=\linewidth]{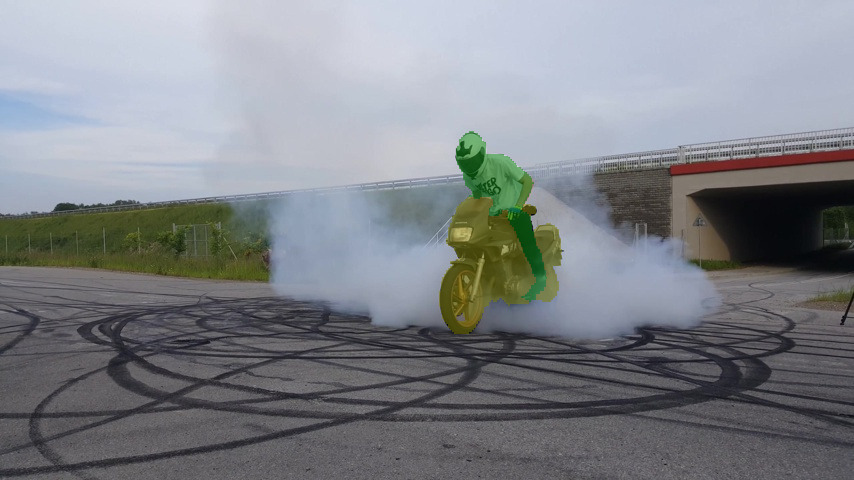}
    \end{subfigure}
    \begin{subfigure}{0.24\textwidth}
      \centering
      \includegraphics[width=\linewidth]{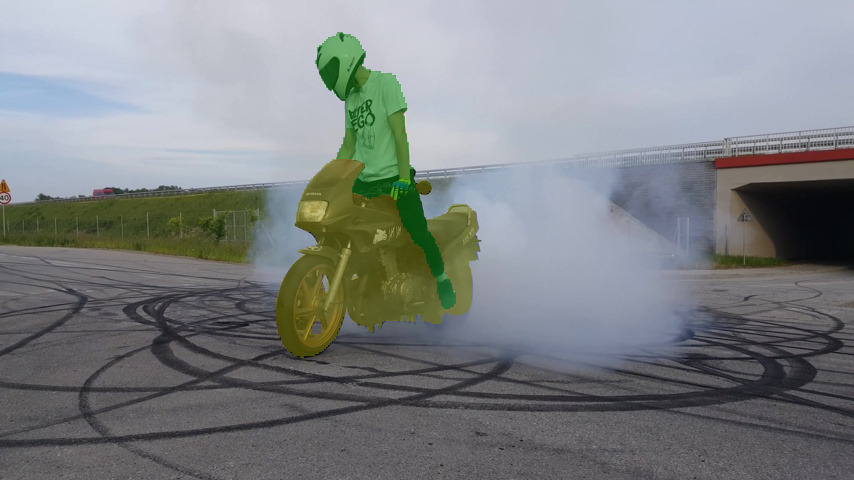}
    \end{subfigure}
    \begin{subfigure}{0.24\textwidth}
      \centering
      \includegraphics[width=\linewidth]{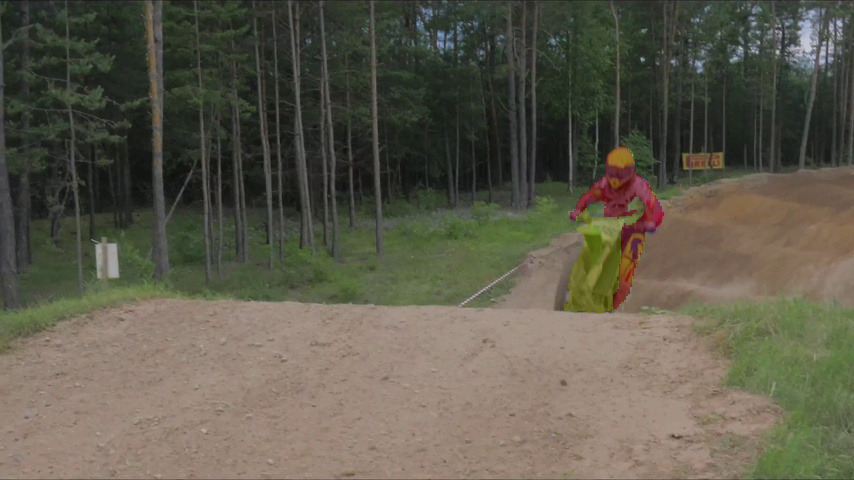}
    \end{subfigure}
    \begin{subfigure}{0.24\textwidth}
      \centering
      \includegraphics[width=\linewidth]{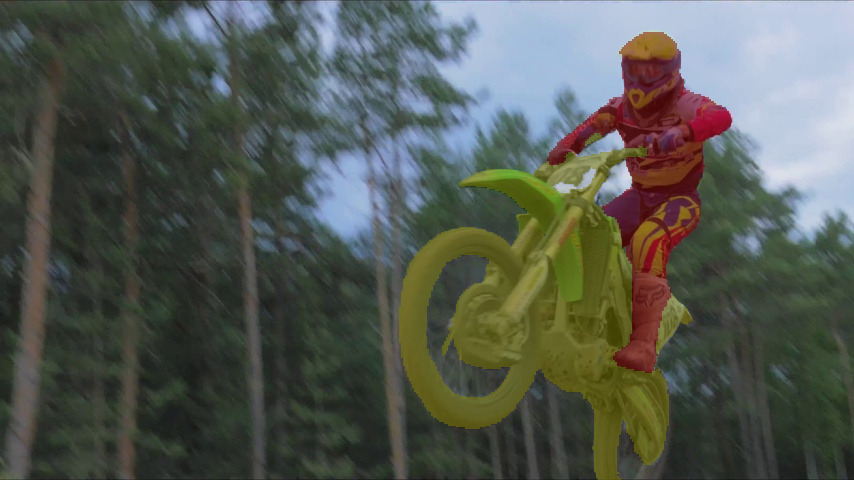}
    \end{subfigure}
    \begin{subfigure}{0.24\textwidth}
      \centering
      \includegraphics[width=\linewidth]{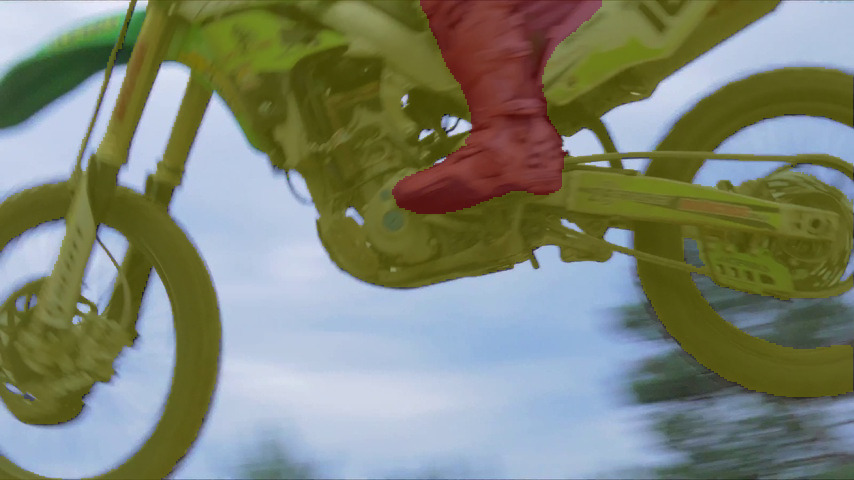}
    \end{subfigure}
    \begin{subfigure}{0.24\textwidth}
      \centering
      \includegraphics[width=\linewidth]{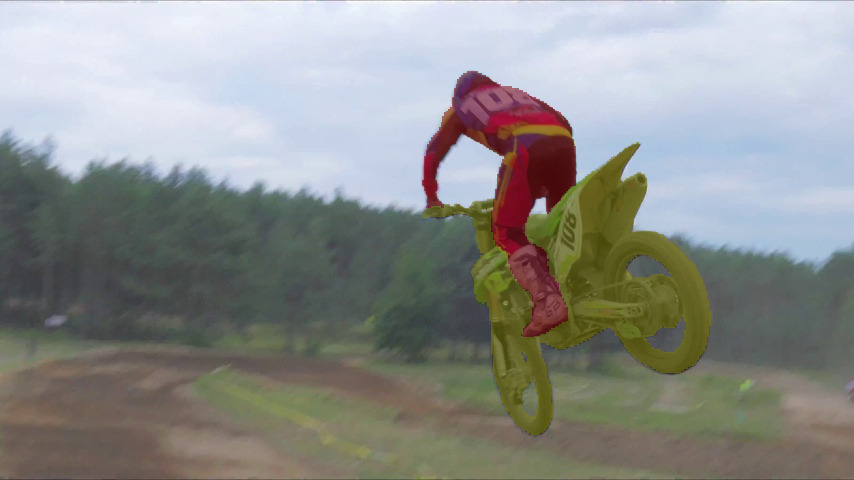}
    \end{subfigure}
    \begin{subfigure}{0.24\textwidth}
      \centering
      \includegraphics[width=\linewidth]{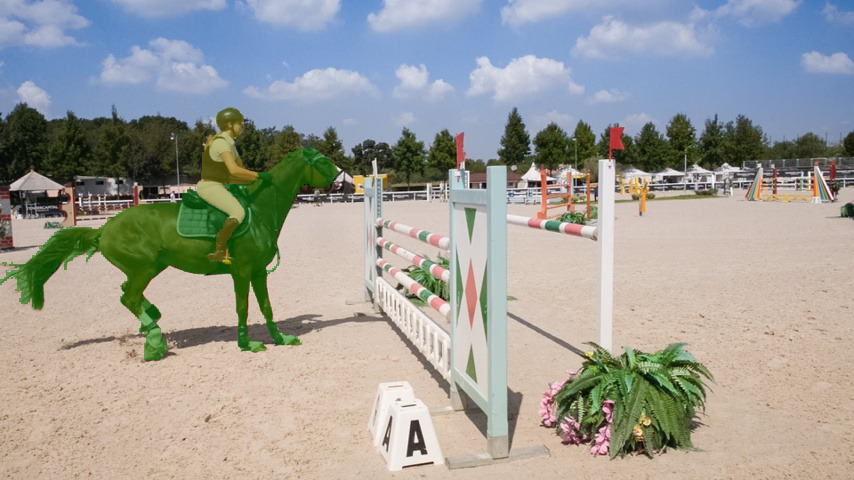}
    \end{subfigure}
    \begin{subfigure}{0.24\textwidth}
      \centering
      \includegraphics[width=\linewidth]{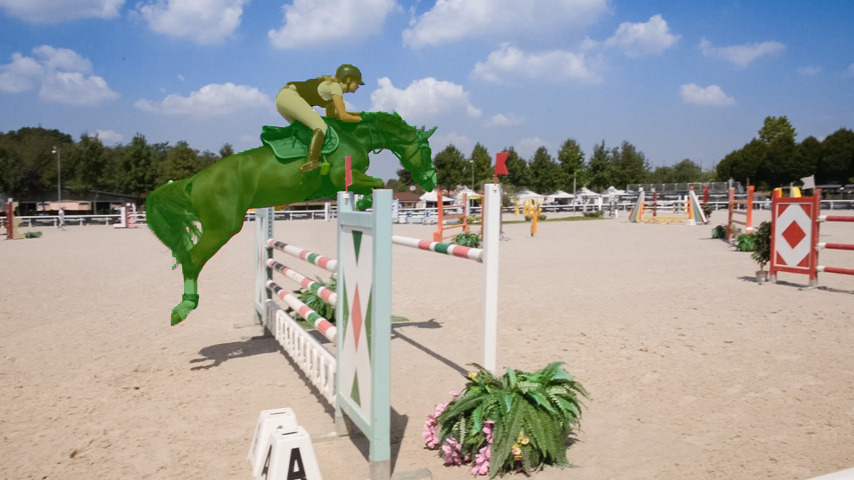}
    \end{subfigure}
    \begin{subfigure}{0.24\textwidth}
      \centering
      \includegraphics[width=\linewidth]{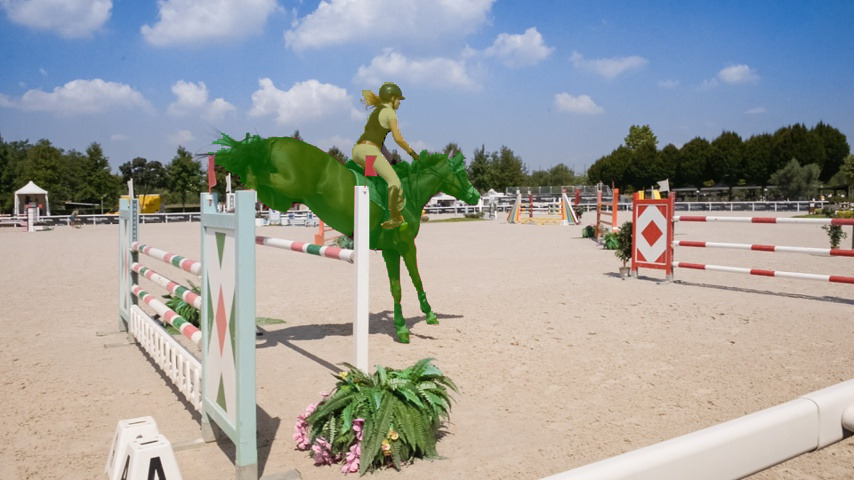}
    \end{subfigure}
    \begin{subfigure}{0.24\textwidth}
      \centering
      \includegraphics[width=\linewidth]{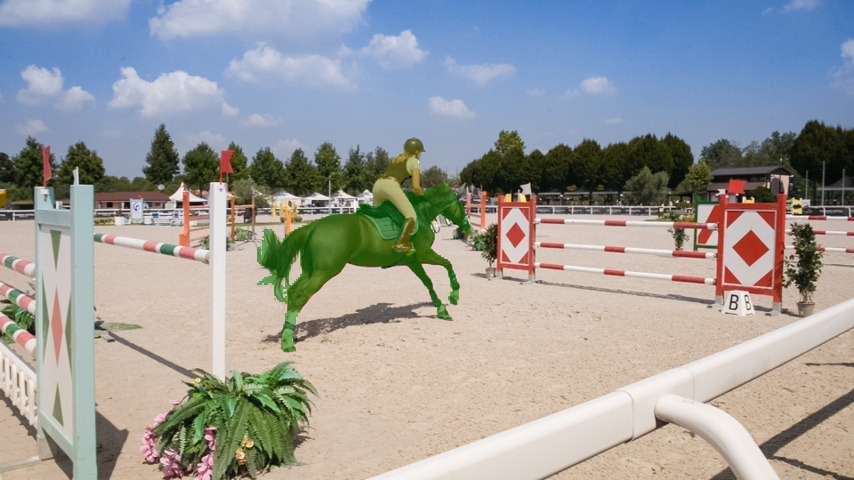}
    \end{subfigure}
    \begin{subfigure}{0.24\textwidth}
      \centering
      \includegraphics[width=\linewidth]{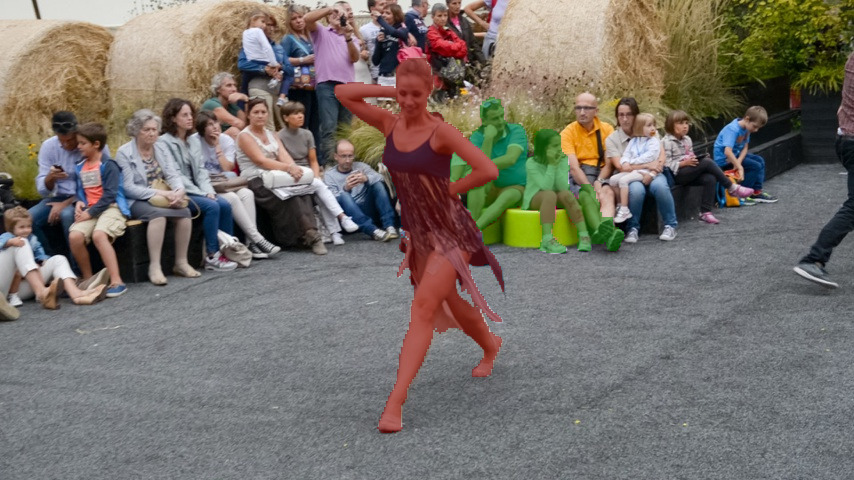}
    \end{subfigure}
    \begin{subfigure}{0.24\textwidth}
      \centering
      \includegraphics[width=\linewidth]{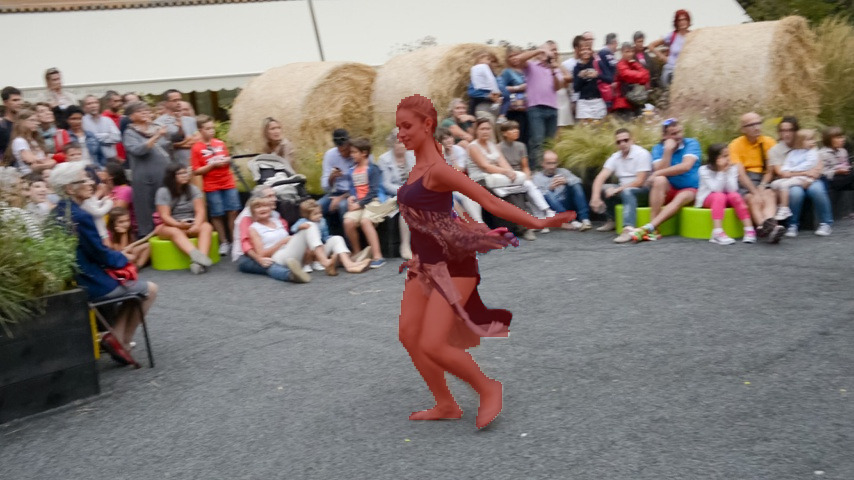}
    \end{subfigure}
    \begin{subfigure}{0.24\textwidth}
      \centering
      \includegraphics[width=\linewidth]{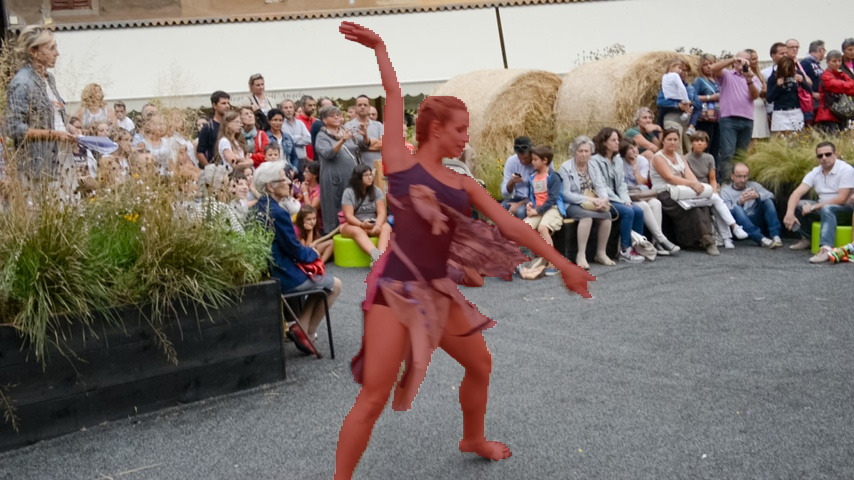}
    \end{subfigure}
    \begin{subfigure}{0.24\textwidth}
      \centering
      \includegraphics[width=\linewidth]{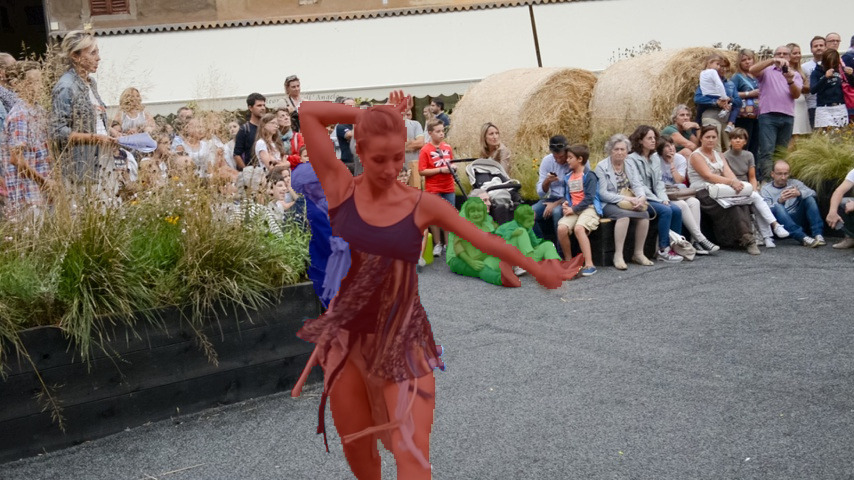}
    \end{subfigure}
    \caption{\label{fig:more_case_davis} Predictions on DAVIS \textit{val} set.}
\end{figure*}

Figure~\ref{fig:more_case_coco1} and~\ref{fig:more_case_coco2} present more visualization of our model's predictions on MS-COCO validation set. Figure~\ref{fig:more_case_cityscapes} and~\ref{fig:more_case_davis} present extra visualization of our model's predictions on Cityscapes and DAVIS validation sets, respectively.

\end{document}